\documentclass[letterpaper,twocolumn,10pt]{article}
\usepackage{usenix}

\usepackage{tikz}
\usepackage{amsmath}
\usepackage{amssymb}
\usepackage{pifont}
\usepackage{tikz}
\usepackage{caption}
\usepackage{subcaption}
\usepackage{booktabs}
\usepackage{multirow}

\usepackage{algorithm}
\usepackage{algorithmic}

\newcommand{\toolns}{\textit{CAD}}
\newcommand{\tool}{\toolns\space}
\newcommand{\datans}{\textit{CADA}}
\newcommand{\data}{\datans\space}
\newcommand{\Tref}[1]{Tab.~\ref{#1}}
\newcommand{\Eref}[1]{Eq.~(\ref{#1})}
\newcommand{\Fref}[1]{Fig.~\ref{#1}}
\newcommand{\Sref}[1]{Sec.~\ref{#1}}

\newcommand{\ie}{\textit{i}.\textit{e}.}
\newcommand{\eg}{\textit{e}.\textit{g}.}
\newcommand{\etal}{\textit{et} \textit{al}.}

\begin{document}

\date{}

\title{Black-Box Adversarial Attack on Vision Language Models\\ for Autonomous Driving}

\author{
{\rm Lu Wang$^{1}$, Tianyuan Zhang$^{1}$, Yang Qu$^{1}$, Siyuan Liang$^{2}$, Yuwei Chen$^{3}$,}\\
{\rm Aishan Liu$^{1}$, Xianglong Liu$^{1}$, Dacheng Tao$^{4}$}\\
$^{1}$Beihang University, $^{2}$National University of Singapore, \\$^{3}$Aviation Industry Development Research Center of China, $^{4}$Nanyang Technological University\and
} 

\maketitle


\begin{abstract}
\label{sec:abstract}

Vision-language models (VLMs) have significantly advanced autonomous driving (AD) by enhancing reasoning capabilities; however, these models remain highly susceptible to adversarial attacks. 
While existing research has explored white-box attacks to some extent, the more practical and challenging black-box scenarios, where neither the model architecture nor parameters are known, remain largely underexplored due to their inherent difficulty. 
In this paper, we take the first step toward designing black-box adversarial attacks specifically targeting VLMs in AD. 
We identify two key challenges for achieving effective black-box attacks in this context: the effectiveness across driving reasoning chains in AD systems and the dynamic nature of driving scenarios. To address this, we propose Cascading Adversarial Disruption (\toolns). It first introduces Decision Chain Disruption, which targets low-level reasoning breakdown by generating and injecting deceptive semantics, ensuring the perturbations remain effective across the entire decision-making chain. Building on this, we present 
Risky Scene Induction, which addresses dynamic adaptation by leveraging a surrogate VLM to understand and construct high-level risky scenarios that are likely to result in critical errors in the current driving contexts. 
Extensive experiments conducted on multiple AD VLMs and benchmarks demonstrate that \tool achieves state-of-the-art attack effectiveness, significantly outperforming existing methods (+13.43\% on average). 
Moreover, we validate its practical applicability through real-world attacks on AD vehicles powered by VLMs, where the route completion rate drops by 61.11\% and the vehicle crashes directly into the obstacle vehicle with adversarial patches. 
Finally, we release the \data dataset, comprising 18,808 adversarial visual-question-answer pairs, to facilitate further evaluation and research in this critical domain. Our codes and dataset will be available after paper's acceptance.
\end{abstract}
\section{Introduction}
\label{sec:introduction}

Recent advancements in vision-language models (VLMs) \cite{alayrac2022flamingo, li2023blip, liu2024llava, zhu2023minigpt} have achieved impressive performance in tasks such as image captioning, question answering, and multimodal reasoning, leading to their rapid development and widespread applications across various domains, including autonomous driving (AD).  
AD operates within highly complex and dynamically evolving environments, where the sophisticated multimodal reasoning capabilities of VLMs could serve as a cognitive backbone, enabling advanced perception, nuanced decision-making, and safer navigation \cite{nie2023reason2drive, marcu2023lingoqa, shao2024lmdrive, sima2023drivelm, mao2023gptdriver, ma2023dolphins}.

\begin{figure}
    \centering
    \includegraphics[width=0.96\linewidth]{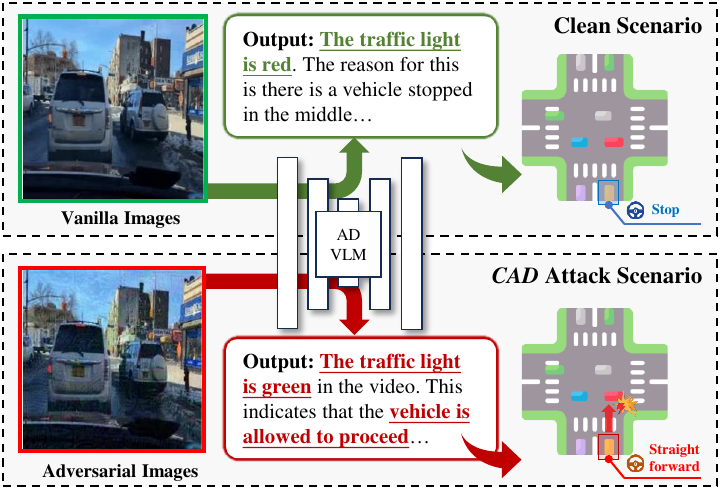}
    \caption{Illustration of our \tool black-box attack. The added visual perturbations cause the model to misinterpret a red traffic light as green. This misjudgment leads to incorrect driving actions, ultimately resulting in a collision.}
    \label{fig:frontpage}
\end{figure}

Despite their remarkable success, numerous studies have highlighted the vulnerability of VLMs to adversarial attacks \cite{wang2024transferable, zhou2023advclip, yin2024vlattack, zhang2022towards,liang2021generate,liang2020efficient,wei2018transferable,liang2022parallel,liang2022large,wang2023diversifying,liu2023x,liu2023improving}, where imperceptible visual perturbations would cause the model to wrong predictions. These vulnerabilities raise significant concerns, especially when such models are deployed in safety-critical AD contexts \cite{sima2023drivelm, zhang2024visual}. Subtle errors at the reasoning stage can cause severe failures, for instance, misinterpreting a situation that demands braking as one requiring acceleration \cite{deng2021deep,gonzalo2020driveless}. Tackling these challenges demands a systematic approach to evaluating the robustness of driving models. Effective attack methods not only uncover latent vulnerabilities but also offer a practical means to assess and enhance model resilience in real-world conditions \cite{GUO2023comprehensive,zhang2024lanevil, zhang2021interpreting, tang2021robustart, liu2021training, liu2023exploring, guo2023towards, liu2023towards,zhang2024module, liang2023badclip, liang2024poisoned}. 

In this domain, the limited existing research predominantly
focused on white-box attacks \cite{zhang2024visual} where attackers have full knowledge of the target models; in contrast, the more practical and challenging black-box attack has been largely underexplored  (\ie, the attackers only possess limited model access). However, it is highly non-trivial to simply extend existing black-box attacks \cite{chen2017zoo} into VLM AD, due to the following challenges: \ding{182} Attack should effectively propagate and amplify errors across the complex driving reasoning chain (perception, prediction, and plan) to achieve coherent and strong threats to overall systems. \ding{183} Attack should account for the dynamic driving context of AD, ensuring the induced errors could compromise driving safety in the given context. 

To address the above problems, we propose the black-box attack, Cascading Adversarial Disruption (\toolns) for VLM AD. Our framework first introduces \textbf{Decision Chain Disruption}, which targets low-level reasoning breakdown by generating and injecting deceptive semantics, ensuring the perturbations remain effective across the entire decision-making chain. 
Building on this, we present \textbf{Risky Scene Induction}, which addresses dynamic adaptation by leveraging a surrogate VLM to understand and construct high-level risky scenarios that are likely to result in critical errors in the current driving contexts.
Finally, by semantic discrepancy maximization as a complementary enhancement and jointly optimizing the adversarial objectives, \tool generates adversarial visual inputs that provoke erroneous behaviors, thereby accomplishing an untargeted attack that undermines the safety and reliability of VLM AD systems.

Extensive experiments are conducted on multiple AD VLMs (Dolphins \cite{ma2023dolphins}, DriveLM \cite{sima2023drivelm} LMDrive \cite{shao2024lmdrive} and general VLMs (InstructBlip \cite{dai2023instructblip}, LLaVA \cite{liu2024llava}, MiniGPTv4 \cite{zhu2023minigpt}, GPT-4o \cite{chatgpt}) in both open-loop and closed-loop environments, where \tool outperforms all baselines by 13.43\% on average. Furthermore, we demonstrate the efficacy of our attack by successfully attacking real AD vehicles driven by the adapted VLMs on the road following task, where the route completion rate drops by 61.11\% and the vehicle crashes directly into the obstacle vehicle with adversarial patches. 
Finally, we present \data dataset, which consists of 18,808 adversarial visual-question-answer pairs to facilitate future research. Our \textbf{contributions} are:

\begin{itemize}
    \item To the best of our knowledge, we propose the \tool Attack, the first black-box adversarial attack specifically tailored for VLM AD based on Risky Scene Induction and Decision Chain Disruption. 
    \item We conduct extensive experiments on multiple benchmarks, as well as the real AD vehicles driven by VLMs, showing the superiority of our attack.
    \item We present the \data dataset, comprising 18,808 adversarial visual-question-answer pairs, to facilitate research.
\end{itemize}

\section{Backgrounds and Preliminaries}
\label{sec:background}

\quad \textbf{Autonomous driving based on VLMs.}
VLMs have demonstrated remarkable performance across a wide range of tasks \cite{zhu2023minigpt, achiam2023gpt, liu2024llava, dai2023instructblip}, and are increasingly explored in AD. 
In particular, VLMs integrate visual and textual information for visual question-answering (VQA) or planning and control tasks. 
In general, an AD VLM $f$ is designed with an architecture that combines a vision encoder $f_{v}$, a large language model (LLM) backbone $f_{l}$, and the post-processing module $f_{p}$, jointly completing the entire task pipeline.  
The inputs consist of video or multi-frame images, represented as $\mathbf{x}^{v} = \{\mathbf{x}^{v}_{1},\ \mathbf{x}^{v}_{2},\ ... ,\ \mathbf{x}^{v}_{n}\}$, where $n$ denotes the maximum number of frames, as well as a textual instruction, expressed as $\mathbf{x}^{t}$, and the set $\left(\mathbf{x}^{v},\ \mathbf{x}^{t}\right)$ collectively forms the input query space $\mathbb{Q}$ of the model $f$.
In AD tasks, $\mathbf{x}^{v}$ is processed by the visual encoder $f_{v}$ to interpret the scene and generate the visual query,
while $\mathbf{x}^{t}$ is passed through the tokenizer $f_{t}$ associated with the LLM backbone $f_{l}$ to produce textual tokens.
After receiving the visual query and text tokens, the LLM backbone performs joint reasoning and produces task-specific output tokens $\textbf{o}$, expressed as follows:
\begin{equation}
\mathbf{o} = f_{l}\left(f_{v}\left(\mathbf{x}^{v}\right),\ f_{t}\left(\mathbf{x}^{t}\right)\right).
\end{equation}
Depending on the specific AD task, $\mathbf{o}$ is then transformed into the corresponding responses $y$ by the post-processing module $f_{p}$.
For VQA tasks, the tokenizer decodes the output tokens $\mathbf{o}$ into corresponding textual answers. While for planning and control tasks, an adapter processes the tokens to predict future waypoints, which are subsequently transformed into control signals, such as brake, throttle, and steer angle. This process can be formally expressed as follows:
\begin{equation}
y = f_{p}\left(\mathbf{o}\right),
\end{equation}
thereby achieving the mapping of \(f: \mathbb{Q} \rightarrow \mathbb{R}\), where $\mathbb{R}$ represents the responses domain related with the AD tasks.
To train AD VLMs, the visual encoder may either be initialized with pre-trained weights from large-scale vision datasets or trained independently with perception tasks, while the entire model is fine-tuned on task-specific datasets for AD applications.

\textbf{Adversarial Attacks on VLMs.} The widespread deployment and impressive performance of VLMs in multimodal question answering and reasoning have raised growing concerns about their adversarial robustness \cite{zhang2024anyattack, zhao2024evaluating}. 
Here, adversarial attacks on VLMs aim to manipulate the model’s final responses by introducing perturbations into the input query $\left(\mathbf{x}^{v}, \mathbf{x}^{t}\right)$, producing an adversarial query \(\mathcal{A}\left(\mathbf{x}^{v},\ \mathbf{x}^{t}\right)\), where $\mathcal{A}\left(\cdot\right)$ denotes the attack function. The goal of \(\mathcal{A}\) is to induce the victim VLM \(f\) to output targeted or undesired responses \({y}^*\) instead of the correct ones. This can be formalized by maximizing the model's log-likelihood of the adversarial responses \({y}^*\), as shown in \Eref{equ:common_def}:

\begin{equation}
\label{equ:common_def}
\text{maximize}  \log p\left(y^* | \mathcal{A}\left(\mathbf{x}_v,\ \mathbf{x}_t\right)\right)
\end{equation}

\noindent where \(p\left(\cdot\right)\) is a probability function.
Depending on whether $\mathcal{A}\left(\cdot\right)$ requires access to model information, attacks are categorized as white-box or black-box.
\section{Threat Model}
\label{sec:thread_model}

\subsection{Problem Definition}

This paper focuses on black-box visual adversarial attacks on AD VLMs, where the adversary lacks access to the model’s internal parameters and can only craft perturbations on the model's visual inputs. 
Specifically, the attack function $\mathcal{A}$ introduced in \Sref{sec:background} becomes $\mathcal{A}\left(\mathbf{x}^{v}\right) = \mathbf{x}^{v} + \mathbf{\delta}$, where $\mathbf{\delta}$ represents the small perturbations applied to the visual domain, while the textual instruction $\mathbf{x}^{t}$ remains unchanged.
We define $\mathbb{R}^{+}$ as the set of model responses that align with the factual context of the scene and comply with human rules, while $\mathbb{R}^{-}$ denotes the set of responses that either contradict reality or violate rules. 
Our attack objective is to generate perturbations under a given budget $\epsilon$ constrained by the $\ell_p$ norm, to craft noise that forces the responses from the victim model $f$ to approach $\mathbb{R}^{-}$, formally expressed as:
\begin{equation}
\label{equ:black-box}
f_{p}\left(f_{l}\left(f_{v}\left(\mathbf{x}^{v}+\delta\right),\ f_{t}\left(\mathbf{x}^{t}\right)\right)\right) \in \mathbb{R}^{-},  \ \text{s.t.} \ \left|\left| \delta \right|\right|_p \le \epsilon,
\end{equation}
where this process relies solely on a few surrogate models and has no access to the victim model $f$.





\subsection{Challenges for AD VLM Attacks}  
Directly applying existing attacks designed for general VLMs to the AD domain presents significant limitations. Specifically, we identify two main \textbf{challenges} as follows:

\textbf{Challenge \ding{182}:} \textit{Attack should remain effective when propagating the driving reasoning chain.} Most AD solutions encompass a reasoning pipeline that integrates perception, prediction, and plan \cite{tian2024drivevlm, jiang2024senna, hu2023uniad, hu2022stp3, casas2021mp3}, where the transition from raw sensory data to higher-level reasoning decisions is inevitable. Conventional visual attacks often focus narrowly on disrupting the perception, such as inducing classification errors, without investigating whether these initial perturbations could remain effective when propagated through subsequent stages or are mitigated by the system's fault-tolerant abilities. To design effective adversarial attacks, it is crucial to account for how errors introduced at the perception stage traverse and accumulate within the reasoning pipeline. By targeting the entire pipeline, adversarial attacks can disrupt reasoning processes in a more sustained and cohesive manner, thereby posing a greater threat to the overall system.

\textbf{Challenge \ding{183}:} \textit{Attack should work for the dynamic driving context of autonomous driving.} Unlike common VLM tasks that work in static context, AD operates within dynamic environments where the determination of safety depends heavily on the specific contextual conditions of the current scenario. For instance, a sudden sharp turn may be the optimal action if a pedestrian unexpectedly enters the lane from the right. However, in standard situations, abrupt maneuvers (\eg sharp turns or hard braking) are typically deemed unsafe. Therefore, the design of adversarial attacks for AD VLMs must account for the contextual dynamics that underpin safe driving behavior, rather than simply triggering isolated extreme actions. Attacks should aim to disrupt decision-making in ways that align with the complexities and situational dependencies inherent to AD, ensuring the induced errors could compromise driving safety within the given context.

\subsection{Adversarial goals}

This paper explores generating visual adversarial perturbations that mislead AD VLMs to wrong responses. Specifically, given an AD VLM \(f\) (consists of $f_{v}$, $f_{t}$, $f_{l}$, and $f_{p}$) that takes video or image sequence \( \mathbf{x}_{v} \) and a textual instruction $\mathbf{x}_{t}$ as input, the attacker’s goal is to induce the model to generate misunderstandings of facts or erroneous driving actions.
We mainly conduct scenario-specific untargeted attacks. A successful attack in this context is achieved when the model's response evaluation score decreases or it induces accident-prone behaviors in the driving agent.

\subsection{Possible attack pathways}

A critical question in adversarial attacks on AD VLMs is whether these attacks are feasible and practical in real-world AD scenarios.
Our approach demonstrates strong applicability in practical AD applications.
On one hand, adversaries can exploit various methods to inject noise into the data collected by sensors during driving. For instance, they may install disruptive devices on sensors to interfere directly with the data captured by the camera, or employ hacking techniques to inject adversarial noise into the images processed by the deployed system model.
On the other hand, physical-world attacks can be realized by optimizing adversarial noise or patches using our proposed objective functions. For example, adversaries could affix adversarial patches onto traffic signs \cite{liu2019perceptual} or embed adversarial noise into roadside billboards \cite{patel2019adaptive}. 


\subsection{Adversary's constraints and capabilities}

In the context of AD, we focus exclusively on black-box attacks, where attackers have no access to any internal information about the victim model (\eg, architecture, gradients, or parameters). 
Querying the model is also impractical in AD applications due to real-time requirements and limited access; therefore, our attacks mainly rely on transferability. 
In addition, attackers are restricted to perturbing visual inputs only, such as directly adding noise in the digital world (\Sref{sec:digital-attack}), and modifying object appearances by pasting patches or modifying textures in the real world (\Sref{sec:close-loop-real}).

\section{Attack Approach}
\label{sec:method}

\begin{figure*}
    \centering
    \includegraphics[width=0.95\linewidth]{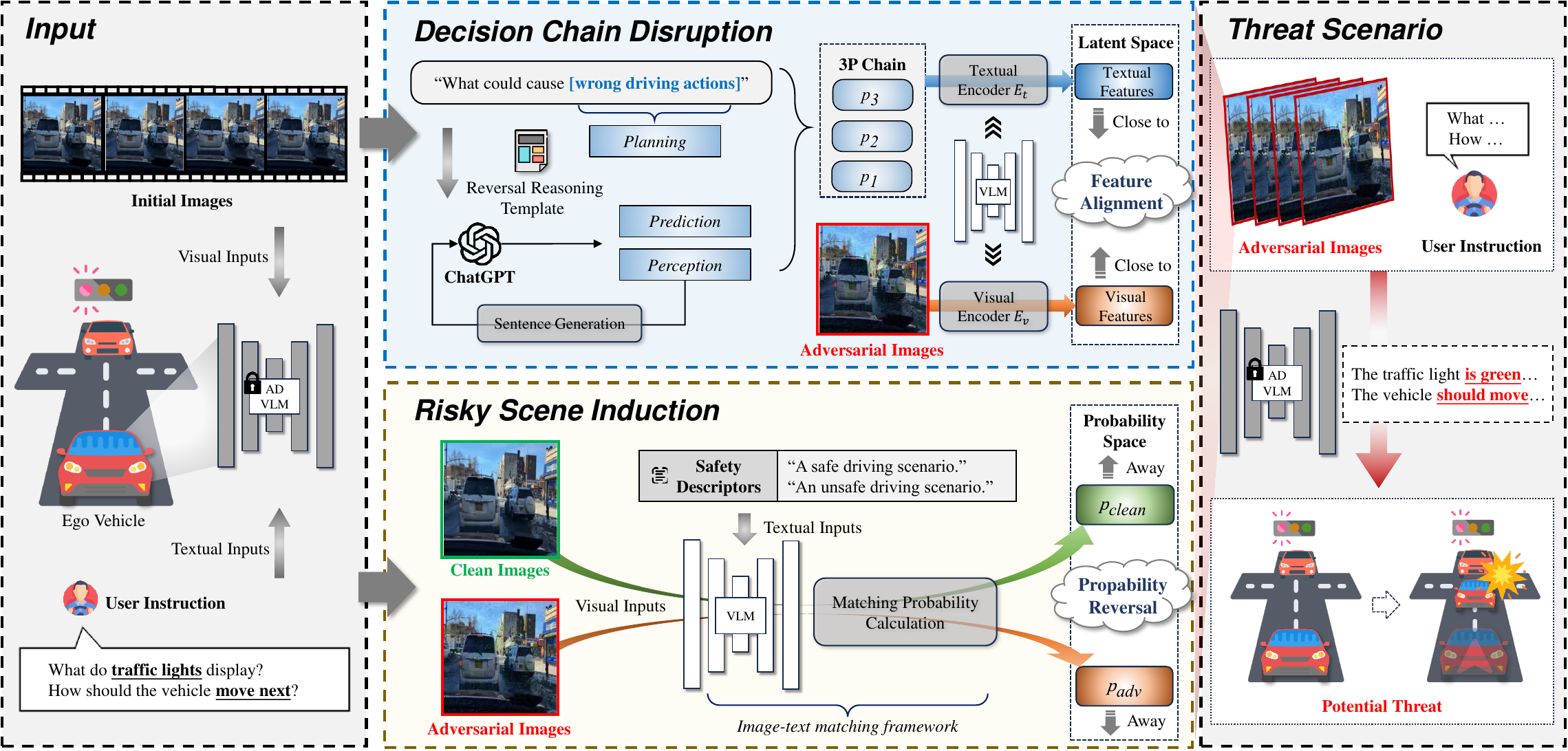}
    \caption{Attack Framework. Our approach introduces Decision Chain Disruption, which targets low-level reasoning breakdown by generating and injecting deceptive semantics.  Building on this, we present Risky Scene Induction, which addresses dynamic adaptation by leveraging a surrogate VLM to understand and construct high-level risky scenarios.}
    \label{fig:framework}
\end{figure*}

As shown in \Fref{fig:framework}, our framework can generate black-box visual adversarial attacks against AD VLMs consisting of two cascading modules: Decision Chain Disruption for \emph{low-level} reasoning breakdown and Risky Scene Induction for \emph{high-level} dynamic adaptation.

\subsection{Decision Chain Disruption}
\label{sec:dcd}

To address Challenge \ding{182}, which emphasizes maintaining attack effectiveness across the driving reasoning chain, we propose a method that specifically targets the perception-prediction-plan pipeline. While conventional VLM attacks often fail to effectively disrupt this reasoning chain, our approach aligns adversarial noise with \emph{low-level} detailed texts that possess coherent yet deceptive semantics, ensuring that disruptions introduced at the perception stage propagate seamlessly through the entire reasoning chain.

Capitalizing on the pivotal role of the perception-prediction-plan reasoning chain in enhancing the effectiveness and reliability of contemporary AD solutions \cite{tian2024drivevlm, jiang2024senna, hu2023uniad, hu2022stp3, casas2021mp3}, we exploit its disruption as a means to generate detailed malicious behaviors against driving VLMs.
We begin by identifying errors in the final stage (\eg, plan) of the corresponding task and then trace backward to infer potential factors that could have contributed to these mistakes in the earlier stages, thereby constructing deceptive texts aligning with real-world scenarios involving unexpected traffic behaviors. 
Formally, our deceptive texts consist of several components corresponding to current and prequel stages. 
We define the expected errors in the form of textual description at the $i_{th}$ stage as $\hat{y}_{i}$. Then starting from each $\hat{y}_{i}$, we iteratively generate the cause-related texts $\hat{y}_{i-1}$ for the preceding stage according to \Eref{eqn:gen_dec_text}:

\begin{equation}
\label{eqn:gen_dec_text}
     \hat{y}_{i-1} = \mathcal{G}\left(T_{R};\ {\textbf{x}}^v,\ \hat{y}_{i}\right),
\end{equation}

\noindent where $\mathcal{G}(\cdot)$ is realized by leveraging an auxiliary VLM (\eg, powerful ChatGPT \cite{chatgpt}) initialized with a \emph{Reversal Reasoning Template} $T_{R}$, which is detailed in the Appendix. 
This process continues backward through each stage, leveraging the powerful logical capabilities of the auxiliary VLM to construct a coherent chain of reasoning that leads to errors.
Ultimately, our deceptive texts $\hat{y}$ can be formalized as a continuous sentence, where each stage's texts are causally connected through reasoning links, as shown in \Eref{eqn:dec_text}:




\begin{equation}
\label{eqn:dec_text}
     \hat{y} = \hat{y}_1 \circ \hat{y}_2 \circ \dots \circ \hat{y}_n,
\end{equation}

\noindent where the symbol \( \circ \) represents the operation of combining the processed text fragments into a single cohesive structure, and $n$ denotes the total stages of the corresponding task. 
Examples of deceptive texts can be found in the Appendix.

After constructing the \emph{low-level} deceptive text $\hat{y}$ in \Eref{eqn:dec_text}, our objective is to optimize the perturbation such that the adversarial images encapsulate the semantics of $\hat{y}$. 
Specifically, by leveraging a pre-trained modality-aligned model $E$ (\eg, CLIP \cite{radford2021clip}) with a visual encoder $E_{v}$ and a textual encoder $E_{t}$, the optimization objective for the adversarial noise $\delta$ is to ensure that the adversarial images and the deceptive texts are as close as possible in the encoded latent space.
To achieve this objective, we define the loss $\mathcal{L}_{l}$ for \emph{low-level} decision chain disruption as:
\begin{equation}
    \label{eqa:loss_bsd}
    \mathcal{L}_{l}\left(\delta;\ \mathbf{x}^{v},\ \hat{y}\right) = 1 - \mathcal{S}\left( E_v\left( \mathbf{x}_{v} + \delta \right),\ E_t\left( \hat{y} \right) \right), 
\end{equation}
\noindent where $ \mathcal{S}(\cdot, \cdot) $ measures the cosine similarity between the visual representation of the adversarial images $\mathbf{x}_{v} + \delta$ encoded by $E_v$ and the textual representation of the deceptive texts $\hat{y}$ encoded by $E_t$. 
By maximizing this similarity, we align the adversarial images with the deceptive semantics of  $\hat{y}$, causing the driving VLMs to misinterpret visual inputs and make erroneous responses. 

\subsection{Risky Scene Induction}
\label{sec:rsi}

While \Sref{sec:dcd} effectively targets low-level reasoning breakdown, it struggles to adapt to the dynamic complexities of real-world driving contexts. To address Challenge \ding{183} of dynamic driving contexts, our method aims to manipulate \emph{high-level} safety assessments, rather than targeting isolated actions, ensuring the attack aligns with the complexities of real-world driving contexts. By disrupting the perceived safety of driving scenarios, we construct holistic risky situations that undermine the model's context-dependent inference.

To achieve this, we leverage a set of opposite descriptors $D$ to represent the safety of a driving scenario. Specifically, we define two complementary categories of scene safety: ``\texttt{A safe driving scenario}'' and ``\texttt{An unsafe driving scenario}''. These descriptors serve as the basis for risky scene induction. 
Subsequently, we integrate an image-text matching framework, where the pre-trained modality-aligned model $E$ in \Sref{sec:dcd} , which includes the visual encoder $E_{v}$ and the textual encoder $E_{t}$, is presented with both the image sequence $\mathbf{x}^{v} = \{\mathbf{x}^{v}_{1},\ \mathbf{x}^{v}_{2},\ ...,\ \mathbf{x}^{v}_{n}\}$ and our predefined textual descriptors $D$ as inputs. 
The model $E$ then generates image feature vectors $E_v\left(\mathbf{x}_v\right)$ and text feature vectors $E_t\left(D\right)$. 
These feature vectors are passed to a downstream matching head $h$, which performs cross-modal matching based on their feature similarity, formally expressed as:

\begin{equation} 
\label{eqn:match} 
h\left(\mathbf{x}^{v},\ D\right) = \text{softmax}\left( \frac{E_v(\mathbf{x}^{v}) @ E_t(D)^T}{\left|\left|E_v(\mathbf{x}^{v})\right|\right|_{2} \cdot \left|\left|E_t(D)\right|\right|_{2}} \right), 
\end{equation}


\noindent where $@$ denotes matrix multiplication and $T$ represents matrix transpose. This yields probabilities that quantify the degree to which a given image aligns with a specific descriptor.
The objective of risky scene induction is to make the matching results of the adversarial image sequence contradict those of the clean image sequence with respect to $D$, thereby constructing a holistic risky scenario.

We feed both the clean and adversarial images into the image-text matching framework in parallel, obtaining the matching results $p_{\text{clean}} = h\left(\mathbf{x}^{v}, D\right)$ and $p_{\text{adv}} = h\left(\mathbf{x}^{v} + \delta, D\right)$ for the clean and adversarial images, respectively. 
Rather than directly associating the adversarial images with the `unsafe' descriptor or merely inverting the prediction probabilities, our goal is to optimize the adversarial images such that the matching outcome contradicts the result $p_{\text{clean}}$, thereby inducing a misanalysis at the holistic scene level. 
Specifically, we initialize a mask $M$ with the same shape as $p_{\text{clean}}$ and $p_{\text{adv}}$ with all zeros and for each image, we set the index corresponding to the class with the lower probability for the clean image to 1, which is formalized in \Eref{eqn:mask}:

\begin{equation} 
\label{eqn:mask} 
M\left[i,\ j\right] = \begin{cases} 
1, & \text{if} \ j = \underset{k}{\arg\min} \left(p_{\text{clean}}\left[i,\ k\right]\right) \\ 
0, & \text{otherwise} \end{cases}, 
\end{equation}


\noindent Then our optimization objective for \emph{high-level} risky scene induction can be expressed as  \Eref{eqn:loss_ssd}:



\begin{equation} 
\label{eqn:loss_ssd} 
\mathcal{L}_h(\delta;\ \mathbf{x}^{v},\ D) = \text{mean}\left(-\log \left(p_{\text{adv}} \odot M \right) + \log\left(p_{\text{adv}} \odot \neg M \right)\right), 
\end{equation}

\noindent where $\odot$ denotes the element-wise multiplication which applies the mask to the respective probabilities, and $\text{mean}\left(\cdot\right)$ denotes the calculation of the average. 

We generate detailed deceptive texts to disrupt the decision chain from a \emph{low-level} perspective and induce a holistic risky scene from a \emph{high-level} perspective. Together, these form the essential foundation and objectives for optimizing adversarial perturbations in the following section.

\subsection{Overall Optimization}
\label{sec:overall}

Building upon the preceding discussions, this section articulates the overall optimization for the adversarial noise $\delta$. 


To complement the first two  objectives further, we introduce the third objective, \ie, semantic discrepancy maximization that maximizes the semantic difference between the clean and adversarial images, thereby reinforcing the overall attack performance. 
The objective introduces perturbations that, while subtle at the pixel level, induce a more significant divergence in the latent space. 
To achieve this, we employ the same $E_{v}$ in \Sref{sec:dcd}, aligned with the text encoder, to map the clean and adversarial images into corresponding visual feature representations. The semantic difference is measured by cosine similarity, leading to the following optimization objective, as the loss $\mathcal{L}_d$ defined in \Eref{eqn:loss_d}:

\begin{equation}
\label{eqn:loss_d}
     \mathcal{L}_d\left(\delta;\ \mathbf{x}^{v}\right) = \mathcal{S}\left({E}_v\left(\mathbf{x}^{v} + \delta\right),\ {E}_v\left(\mathbf{x}^{v}\right)\right),
\end{equation}


\noindent where $ \mathcal{S}(\cdot, \cdot) $ measures the cosine similarity. Thanks to the exceptional capabilities of pre-trained modality-aligned models, we can introduce disruptive semantics in the visual modality to address both challenges without needing access to the victim model, and this allows for supplementation and enhancement through semantic discrepancy.

The final optimization objective integrates $\mathcal{L}_l$ for decision chain disruption , $\mathcal{L}_h$ for risky scene induction,  and $\mathcal{L}_d$ for semantic discrepancy maximization, formalized as \Eref{eqn:overall_loss}:

\begin{equation}
\label{eqn:overall_loss}
\text{minimize}\ \mathcal{L}_{adv}(\delta) = \alpha \cdot \mathcal{L}_{l} + \beta \cdot \mathcal{L}_{h} + \gamma \cdot \mathcal{L}_{d},
\end{equation}

\noindent where $\alpha$, $\beta$ and $\gamma$ are hyperparameters that control the relative weight of each loss term in the overall optimization. We utilize a gradient-based optimization framework \cite{PGD} to iteratively refine the adversarial noise $\delta$. Following each update, we project the noise back into the permissible perturbation budget $\epsilon$, thereby ensuring that the adversarial noise remains subtle and imperceptible while concurrently minimizing the associated loss functions.  
To further enhance the optimization process, we incorporate a momentum mechanism \cite{dong2018mifgsm}, which smooths the update trajectory and mitigates the risk of convergence to suboptimal local minima.
The pseudo-algorithm code of our \tool can be found in the Appendix.

\section{Experiments and Evaluations}
\label{sec:exp}

\subsection{Experimental Setup}
\label{sec:setup}

\quad \textbf{Target Models.} 
In our experiments, we assess three state-of-the-art open-source AD VLMs, addressing both Visual Question Answering (VQA) and planning control tasks. These include Dolphins \cite{ma2023dolphins}, built on OpenFlamingo \cite{alayrac2022flamingo}, for driving-related dialogues; DriveLM \cite{sima2023drivelm}, a LLaMA-based GVQA model even capable of coordinate-level recognition; and LMDrive \cite{shao2024lmdrive}, a LLaMA-based agent for closed-loop control in the Carla simulator \cite{dosovitskiy2017carla}. In addition, we also evaluate the robustness of general VLMs adapted for AD tasks, including InstructBlip \cite{dai2023instructblip}, LLaVA \cite{liu2024llava}, MiniGPTv4 \cite{zhu2023minigpt} and GPT-4o \cite{chatgpt}.

\textbf{Datasets.} We evaluate the three driving-specific models on their respective benchmarks. The Dolphins Benchmark \cite{ma2023dolphins} and DriveLM-NuScenes \cite{sima2023drivelm} are primarily composed of question-answering data. For LMDrive, we assess closed-loop driving control on the LangAuto Benchmark-short \cite{shao2024lmdrive}. As for general VLMs, we adapt them to the well-established Dolphins Benchmark for robustness evaluation.

\textbf{Evaluation Metrics.} 
For Dolphins Benchmark, we adopt their original evaluation metrics \cite{ma2023dolphins}, where the final score is the average score across six categories of tasks, where each task is evaluated by Accuracy, Language Score and GPT Score. 
For DriveLM, we extend the metrics used in \cite{sima2023drivelm}. The original evaluation approach employs four distinct metrics across different question types: Accuracy, ChatGPT Score, Match Score, and Language Score. To provide a more comprehensive evaluation, we also introduce the GPT Score for the subset data which employs Language Score, represented by Lan2GPT, and the final evaluation is based on the average of all five metrics. 
For LMDrive, we consider three key metrics proposed in the CARLA Leaderboard \cite{CarlaLeaderboard}: route completion (RC), infraction score (IS), and driving score (DS). \emph{For all these metrics, the lower the values ($\textcolor{blue}{\downarrow}$) the better the attacks.} Further details on these metrics can be found in the Appendix.

\textbf{Compared Attacks.} 
Given the absence of black-box attacks specifically targeting driving VLMs, we adopt general adversarial attack methods as comparison baselines. We select three classic adversarial attack methods, including gradient-based FGSM \cite{goodfellow2014fgsm} and PGD \cite{PGD} attacks, which employ model transfer techniques for black-box attacks, as well as the query-based ZOO \cite{chen2017zoo}, employing both Adam and Newton optimization. 
Additionally, we consider three effective black-box attacks on visual modalities of general VLMs, namely AdvClip \cite{zhou2023advclip}, AttackVLM \cite{zhao2024evaluating}, and AnyAttack \cite{zhang2024anyattack}, as well as two attacks targeting both modalities, SGA \cite{lu2023sga} and VLPAttack \cite{gao2025vlpattack}. We only utilize these two methods to generate noise on the visual inputs to ensure consistency.
We also conduct a comparison with the black-box transferability of the only white-box attack ADvLM \cite{zhang2024visual} targeting AD VLMs.

\textbf{Implementation Details.} In our attacks, we set the hyperparameters $\alpha$, $\beta$, and $\gamma$ to 0.75, 0.05, and 0.75, respectively, based on the results from ablation experiments in \Sref{sec:ablation}. We define the perturbation budget $\epsilon$ as 0.1 under the $\ell_\infty$ norm and set the number of iterations $N$ to 160, keeping the comparison methods consistent. We utilize GPT-4o \cite{chatgpt} as the auxiliary VLM for Decision Chain Disruption in \Sref{sec:dcd} as well as the GPT Score evaluation, while CLIP \cite{radford2021clip} serves as the pre-trained modality-aligned model $E$ in \Sref{sec:method}.
All code is implemented in PyTorch, and experiments are conducted on an NVIDIA A800-SXM4-80GB GPU cluster. 

\subsection{Digital World Experiments}
\label{sec:digital-attack}
We first perform digital world attacks in both the open-loop (\ie, the model operates independently without feedback from the environment) and closed-loop (\ie, the model interacts with the environment and adjusts its behavior based on real-time feedback) evaluation setups.

\begin{table}[!t]
\centering
\caption{Evaluation results of open-loop results in the digital world. \textbf{Bold text} indicates the method with the strongest attack effect in each column. \colorbox[gray]{0.9}{Gray cells} represent comprehensive evaluation metrics. For all metrics, lower values ($\textcolor{blue}{\downarrow}$) indicate stronger attack performance.}
\label{tab:open-digital}
\subfloat[Results on Dolphins \cite{ma2023dolphins}. The final score is the average of scores from previous 6 tasks. ``Desc.'' indicates the detailed description task.]{
\label{tab:dolphin}

\resizebox{\linewidth}{!}{
\centering
\setlength{\tabcolsep}{1.8pt}
\centering
\begin{tabular}{@{}ccccccc>{\columncolor{gray!20}}cc@{}}
\toprule
\textbf{Model} & Weather$\textcolor{blue}{\downarrow}$ & Traffic$\textcolor{blue}{\downarrow}$ & Time$\textcolor{blue}{\downarrow}$ & Scene$\textcolor{blue}{\downarrow}$ & Object$\textcolor{blue}{\downarrow}$ & Desc.$\textcolor{blue}{\downarrow}$ & Final Score$\textcolor{blue}{\downarrow}$ \\ \midrule
Origin                      & 51.14 & 54.91 & 44.12 & 45.70 & 30.24 & 43.17 & 44.88 \footnotesize{($\pm{}$0.97)} \\ \midrule
FGSM\cite{goodfellow2014fgsm} & 71.59 & 48.34 & 36.85 & 46.32 & 29.35 & 34.80 & 44.54 \footnotesize{($\pm{}$1.22)} \\
PGD\cite{PGD}  & 48.89 & 51.52 & 38.51 & 43.38 & 32.06 & 34.55 & 41.49 \footnotesize{($\pm{}$0.99)} \\
ZOO-Adam\cite{chen2017zoo}    & 46.91 & 50.17 & 51.52 & 42.46 & 33.31 & 36.03 & 43.40 \footnotesize{($\pm{}$1.12)} \\
ZOO-Newt\cite{chen2017zoo}     & 49.68 & 50.76 & 36.21 & 44.71 & 37.12 & 35.35 & 42.30 \footnotesize{($\pm{}$1.67)} \\
AdvClip\cite{zhou2023advclip}                   & 49.21 & 52.09 & 39.87 & 44.28 & 26.81 & 35.74 & 41.33 \footnotesize{($\pm{}$1.31)} \\
AttackVLM\cite{zhao2024evaluating}                & 48.33 & 51.48 & 38.11 & 41.26 & 28.48 & 38.42 & 41.01 \footnotesize{($\pm{}$1.09)} \\
AnyAttack\cite{zhang2024anyattack}                & 40.01 & 49.44 & 37.47 & 39.46 & 26.63 & 37.51 & 38.42 \footnotesize{($\pm{}$1.25)} \\
SGA\cite{lu2023sga}                      & 48.89 & 55.62 & 39.17 & 44.92 & 28.94 & 42.01 & 43.26 \footnotesize{($\pm{}$1.15)} \\

VLPAttack\cite{gao2025vlpattack}                 & 50.35 & 52.37 & 38.65 & 42.92 & 26.54 & 36.16 & 41.16 \footnotesize{($\pm{}$0.89)} \\
ADvLM\cite{zhang2024visual}                 & 46.22 & 50.60  & \textbf{35.41} & \textbf{37.56}  & 23.94  & 37.47 & 38.53 \footnotesize{($\pm{}$0.97)} \\
\midrule
\toolns & \textbf{36.75} & \textbf{47.13} & 36.30 & 38.18 & \textbf{16.67} & \textbf{32.99} & \textbf{34.67} \footnotesize{($\pm{}$0.98)} \\ \bottomrule
\end{tabular}

}
}


\vspace{0.2cm}
\subfloat[Results on DriveLM \cite{sima2023drivelm}. The final score is the average of the previous five metrics. ``Acc.'' indicates the Accuracy and ``Lang.'' indicates the Language Score.]{
\label{tab:drivelm}

\resizebox{\linewidth}{!}{
\centering
\setlength{\tabcolsep}{1.8pt}
\begin{tabular}{@{}cccccc>{\columncolor{gray!20}}cc@{}}
\toprule
\textbf{Method}  & Chatgpt$\textcolor{blue}{\downarrow}$ & Acc.$\textcolor{blue}{\downarrow}$ & Match$\textcolor{blue}{\downarrow}$ & Lan2GPT$\textcolor{blue}{\downarrow}$ & Lang.$\textcolor{blue}{\downarrow}$ & Final Score$\textcolor{blue}{\downarrow}$ \\ \midrule
Origin           & 69.77 & \textbf{70.00} & 38.31 & 41.54 & 48.16 & 53.55 \footnotesize{($\pm{}$0.72)} \\ \midrule
FGSM\cite{goodfellow2014fgsm}      & 67.36 & \textbf{70.00} & 27.31 & 37.05 & 47.12 & 49.77 \footnotesize{($\pm{}$0.68)} \\
PGD\cite{PGD}       & 67.63 & 71.43 & 22.33 & 38.08 & 46.69 & 49.23 \footnotesize{($\pm{}$0.64)} \\
ZOO-Adam\cite{chen2017zoo} & 65.90 & \textbf{70.00} & 26.72 & 34.54 & 48.19 & 49.07 \footnotesize{($\pm{}$0.70)} \\
ZOO-Newt\cite{chen2017zoo} & \textbf{65.51} & \textbf{70.00} & 25.47 & 34.36 & 48.22 & 48.71 \footnotesize{($\pm{}$0.65)} \\
AdvClip\cite{zhou2023advclip}    & 66.78 & 71.43 & 23.08 & 37.72 & 46.69 & 49.14 \footnotesize{($\pm{}$0.66)} \\
AttackVLM\cite{zhao2024evaluating} & 66.41 & \textbf{70.00} & 26.06 & 35.64 & 47.64 & 49.15 \footnotesize{($\pm{}$0.63)} \\
AnyAttack\cite{zhang2024anyattack} & 66.67 & \textbf{70.00} & 22.61 & 37.18 & 46.67 & 48.62 \footnotesize{($\pm{}$0.69)} \\
SGA\cite{lu2023sga}            & 69.78 & 71.43 & 23.36 & 36.23 & 46.79 & 49.52 \footnotesize{($\pm{}$0.61)} \\
VLPAttack\cite{gao2025vlpattack}     & 68.90 & 71.43 & 22.78 & 35.95 & 46.45 & 49.10 \footnotesize{($\pm{}$0.67)} \\
ADvLM\cite{zhang2024visual}                 & 66.41 & \textbf{70.00} & 26.97 & 35.10 & 48.18 & 49.33 \footnotesize{($\pm{}$0.60)} \\
 \midrule
\toolns    & 68.87 & \textbf{70.00} & \textbf{19.58} & \textbf{32.31} & \textbf{40.86} & \textbf{46.32} \footnotesize{($\pm{}$0.60)} \\ \bottomrule
\end{tabular}
}
}
\end{table}

\begin{table}[!t]
\caption{Evaluation results on general VLMs on Dolphins Benchmark. We report the final score for each model and the breakdown of task scores is detailed in the Appendix. \textbf{Bold text} indicates the method with the strongest attack effect in each column. For all metrics, lower values ($\textcolor{blue}{\downarrow}$) indicate stronger attack performance.}
\label{tab:general}
\resizebox{\linewidth}{!}{
\centering
\setlength{\tabcolsep}{0.5pt}
\begin{tabular}{@{}cccccc@{}}
\toprule
\textbf{Method} & InstructBlip\cite{dai2023instructblip}$\textcolor{blue}{\downarrow}$  & LLaVA\cite{liu2024llava}$\textcolor{blue}{\downarrow}$  & MiniGPTv4\cite{zhu2023minigpt}$\textcolor{blue}{\downarrow}$ & GPT-4o\cite{chatgpt}$\textcolor{blue}{\downarrow}$\\ \midrule
Origin      & 41.69 \footnotesize{($\pm{}$1.60)} & 45.50 \footnotesize{($\pm{}$1.45)} & 30.21 \footnotesize{($\pm{}$1.12)} & 47.61 \footnotesize{($\pm{}$1.80)}\\ \midrule
FGSM\cite{goodfellow2014fgsm}     & 36.58 \footnotesize{($\pm{}$1.35)} & 45.92 \footnotesize{($\pm{}$1.55)} & 30.13 \footnotesize{($\pm{}$1.08)} & 46.26 \footnotesize{($\pm{}$1.68)}\\
PGD\cite{PGD}      & 36.94 \footnotesize{($\pm{}$1.27)} & 45.27 \footnotesize{($\pm{}$1.49)} & 29.82 \footnotesize{($\pm{}$1.02)} & 45.93 \footnotesize{($\pm{}$1.63)}\\
ZOO-Adam\cite{chen2017zoo} & 36.86 \footnotesize{($\pm{}$1.18)} & 44.62 \footnotesize{($\pm{}$1.43)} & 29.84 \footnotesize{($\pm{}$1.04)} & 47.30 \footnotesize{($\pm{}$1.56)}\\
ZOO-Newt\cite{chen2017zoo} & 36.72 \footnotesize{($\pm{}$1.22)} & 45.49 \footnotesize{($\pm{}$1.51)} & 29.32 \footnotesize{($\pm{}$1.06)} & 47.10 \footnotesize{($\pm{}$1.62)}\\
AdvClip\cite{zhou2023advclip}    & 37.31 \footnotesize{($\pm{}$1.38)} & 45.35 \footnotesize{($\pm{}$1.57)} & 29.71 \footnotesize{($\pm{}$1.13)} & 47.25 \footnotesize{($\pm{}$1.72)}\\
AttackVLM\cite{zhao2024evaluating}   & 37.14 \footnotesize{($\pm{}$1.32)} & 44.87 \footnotesize{($\pm{}$1.53)} & 29.93 \footnotesize{($\pm{}$1.06)} & 45.68 \footnotesize{($\pm{}$1.64)}\\
AnyAttack\cite{zhang2024anyattack} & 33.53 \footnotesize{($\pm{}$1.22)} & 39.09 \footnotesize{($\pm{}$1.30)} & 28.96 \footnotesize{($\pm{}$1.10)} & 39.57 \footnotesize{($\pm{}$1.42)}\\
SGA\cite{lu2023sga}        & 35.85 \footnotesize{($\pm{}$1.28)} & 44.79 \footnotesize{($\pm{}$1.48)} & 30.12 \footnotesize{($\pm{}$1.00)} & 47.35 \footnotesize{($\pm{}$1.65)}\\
VLPAttack\cite{gao2025vlpattack}  & 36.14 \footnotesize{($\pm{}$1.21)} & 45.53 \footnotesize{($\pm{}$1.52)} & 30.02 \footnotesize{($\pm{}$1.08)} & 46.81 \footnotesize{($\pm{}$1.70)}\\ 
ADvLM\cite{zhang2024visual} & 36.31 \footnotesize{($\pm{}$1.19)} & 44.97 \footnotesize{($\pm{}$1.75)} & 30.05 \footnotesize{($\pm{}$0.98)} & 45.50 \footnotesize{($\pm{}$1.51)}\\ 
\midrule
\toolns      & \textbf{32.34} \footnotesize{($\pm{}$1.03)} & \textbf{36.90} \footnotesize{($\pm{}$1.20)} & \textbf{26.82} \footnotesize{($\pm{}$1.05)} & \textbf{33.90} \footnotesize{($\pm{}$1.30)} \\  \bottomrule
\end{tabular}
}
\end{table}

\textbf{Open-loop Evaluation.} We first evaluate our \tool Attack in open-loop AD tasks, including driving-specific models, \ie, Dolphins \cite{ma2023dolphins} and DriveLM \cite{sima2023drivelm}, and adapted general VLMs \cite{dai2023instructblip, zhu2023minigpt, liu2024llava, chatgpt}. For transfer-based attacks (\eg, FGSM, PGD), we respectively use Dolphins or DriveLM as the surrogate model to generate noise and then transfer on another; for other black-box attacks, they directly query the target models. Here, we run each experiment three times and calculate the average results. As shown in \Tref{tab:open-digital}, we can identify:

\ding{182} In comparison to all baseline methods, our \tool Attack demonstrates the most powerful adversarial performance, yielding an average reduction of 19.60\% in the final score across two driving-specific VLMs and four general VLMs. 
Remarkably, it also achieves successful attacks on the widely-used commercial model, GPT-4o \cite{chatgpt}. 
Instances of erroneous inferences produced by the attacked models can be found in the Appendix. 
Notably, 
our \tool induces  significant logical inconsistencies in the driving model's reasoning and results in serious planning errors.

\ding{183} Among all attack methods, the traditional adversarial attacks (such as PGD \cite{PGD}) exhibited the weakest performance, inducing only a minimal drop of 4.68\%. 
AnyAttack \cite{zhang2024anyattack} proved to be the most effective among compared methods, resulting in an average performance degradation of 13.05\%. We attribute the success of this approach to its generation-based attack mechanism, where the perturbation itself is an image that inherently conveys explicit information. 
Furthermore, the visual component of the dual-modal attacks \cite{lu2023sga, gao2025vlpattack} still performs slightly worse than AnyAttack \cite{zhang2024anyattack} in terms of effectiveness, with an average performance degradation of 4.98\%. 
Ultimately, our \tool still outperforms the transferred form of the white-box attack ADvLM \cite{zhang2024visual} by 12.76\%.

\ding{184} The general VLMs are evaluated on the Dolphins Benchmark due to the openness of its question-answering data. We report the final score for each model, and the complete breakdown of the task scores can be found in the Appendix. As the results shown in \Tref{tab:general}, our \tool demonstrates excellent attack effectiveness on general VLMs used for AD tasks, achieving an average performance drop of 20.34\%. This highlights the strong generalization ability across different models. Both LLaVA \cite{liu2024llava} and GPT-4o \cite{chatgpt} exhibit powerful reasoning capabilities for AD tasks, and even without domain-specific fine-tuning, their final scores surpass those of the original Dolphins \cite{ma2023dolphins}. However, despite their inherent strength, they still lack robustness against carefully crafted adversarial attacks.

\ding{185} Upon a comprehensive analysis of the evaluated models, we find that Dolphins \cite{ma2023dolphins} and GPT-4o \cite{chatgpt} exhibit the weakest adversarial robustness, with an average performance drop of 22.75\% and 28.80\% under our \tool Attack. In contrast, LLaVA \cite{liu2024llava} and miniGPTv4 \cite{zhu2023minigpt} demonstrate relatively stronger robustness, with an average performance decline of 13.50\% and 11.22\%. Additionally, a potential pattern emerges, indicating that models with inherently higher performance tend to exhibit relatively weaker adversarial robustness. 


\begin{table}[!t]
\centering
\caption{Evaluation results of close-loop results in the digital world. \textbf{Bold text} indicates the method with the strongest attack effect in each column. \colorbox[gray]{0.9}{Gray cells} represent the main metric of the leaderboard, serving as the product between RC and IS. For all metrics, lower values ($\textcolor{blue}{\downarrow}$) indicate stronger attack performance.}
\label{tab:close-digital}
\setlength{\tabcolsep}{15pt}
\resizebox{\linewidth}{!}{
\centering
\begin{tabular}{@{}c>{\columncolor{gray!20}}cccc@{}}
\toprule
\textbf{Method} & DS$\textcolor{blue}{\downarrow}$ & RC$\textcolor{blue}{\downarrow}$ & IS$\textcolor{blue}{\downarrow}$  \\ \midrule
Origin          & 49.29 \footnotesize{($\pm{}$0.53)} & 59.32 \footnotesize{($\pm{}$6.69)} & 0.86 \footnotesize{($\pm{}$0.07)}\\ \midrule
FGSM\cite{goodfellow2014fgsm} & 45.54 \footnotesize{($\pm{}$3.66)} & 58.37 \footnotesize{($\pm{}$3.90)} & 0.82 \footnotesize{($\pm{}$0.04)}\\
PGD\cite{PGD}  & 47.53 \footnotesize{($\pm{}$0.99)} & 56.06 \footnotesize{($\pm{}$1.61)} & 0.87 \footnotesize{($\pm{}$0.02)}\\
ZOO-Adam\cite{chen2017zoo}     & 44.09 \footnotesize{($\pm{}$4.79)} & \textbf{51.79} \footnotesize{($\pm{}$8.11)} & 0.89 \footnotesize{($\pm{}$0.05)}\\
ZOO-Newt\cite{chen2017zoo}     & 47.12 \footnotesize{($\pm{}$2.06)} & 56.65\footnotesize{($\pm{}$2.38)} & 0.85 \footnotesize{($\pm{}$0.01)}\\
AdvClip\cite{zhou2023advclip}       & 47.43 \footnotesize{($\pm{}$0.55)} & 58.50 \footnotesize{($\pm{}$2.12)} & 0.83 \footnotesize{($\pm{}$0.03)} \\
AttackVLM\cite{zhao2024evaluating}       & 46.74 \footnotesize{($\pm{}$3.56)} & 55.00 \footnotesize{($\pm{}$5.49)} & 0.86 \footnotesize{($\pm{}$0.00)}\\
AnyAttack\cite{zhang2024anyattack}     & 46.91 \footnotesize{($\pm{}$2.49)} & 59.67 \footnotesize{($\pm{}$3.77)} & 0.81 \footnotesize{($\pm{}$0.03)}  \\
SGA\cite{lu2023sga}            & 42.92 \footnotesize{($\pm{}$2.16)} & 54.89 \footnotesize{($\pm{}$2.38)} & 0.82 \footnotesize{($\pm{}$0.01)}\\
VLPAttack\cite{gao2025vlpattack}     & 44.22 \footnotesize{($\pm{}$4.04)} & 58.99 \footnotesize{($\pm{}$2.03)} & 0.78 \footnotesize{($\pm{}$0.04)}\\ 
ADvLM\cite{zhang2024visual}     & 46.99 \footnotesize{($\pm{}$3.85)} & 59.99 \footnotesize{($\pm{}$0.98)} & 0.82 \footnotesize{($\pm{}$0.02)}\\
\midrule
\toolns        & \textbf{39.99} \footnotesize{($\pm{}$0.85)} & 54.59 \footnotesize{($\pm{}$2.10)}  & \textbf{0.77} \footnotesize{($\pm{}$0.02)} \\ \bottomrule
\end{tabular}
}
\end{table}

\begin{figure}[t]
    \centering
    \begin{subfigure}{0.98\linewidth}
        \centering
        \includegraphics[width=1.0\linewidth]{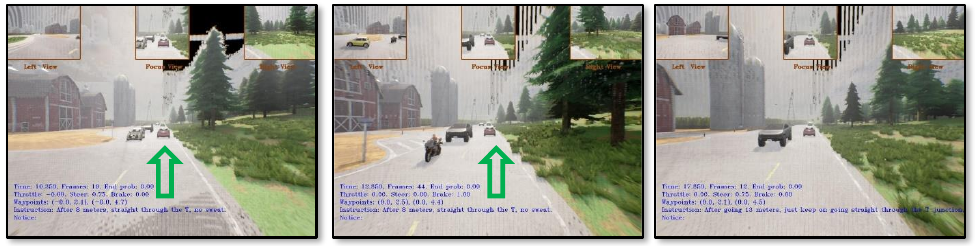}
        \caption{LMDirve without attack: go straight.}
        \label{subfig:carla-clean}
    \end{subfigure}
    
    \begin{subfigure}{0.98\linewidth}
        \centering
        \includegraphics[width=1.0\linewidth]{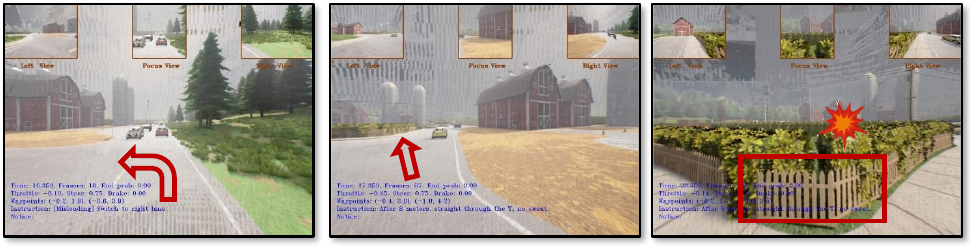}
        \caption{LMDirve under \tool attack: turn left and collide with fence.}
        \label{subfig:carla-attack}
    \end{subfigure}
    
    \caption{Experimental results of LMDrive in the closed-loop simulator, comparing the driving operations of LMDrive at the same location on the same route with and without attack.}
    \label{fig:carla}
\end{figure}

\textbf{Closed-loop Evaluation in Simulator.} For the closed-loop experiments, we utilize the open-source CARLA simulator \cite{dosovitskiy2017carla} to build the simulation environment and employ the state-of-the-art closed-loop driving VLM, LMDrive \cite{shao2024lmdrive}, from the CARLA leaderboard as the driving agent. The attack configuration remains consistent with the open-loop experiments. Specifically, our attack evaluation follows this process: \ding{182} start the CARLA server with a  version of 0.9.10.1, \ding{183} launch the CARLA leaderboard with the LMDrive agent, \ding{184} conduct \tool Attack on the camera images at a frequency of every 50 frames, 
\ding{185} evaluate the driving score. Due to the variability of traffic flow in the simulator, the results on the LangAuto Benchmark show instability. Therefore, we run each set of experiments three times and report the average results, consistent with the settings in \cite{shao2024lmdrive}.

As shown in \Tref{tab:close-digital}, our \tool demonstrates high effectiveness in the closed-loop experiments, resulting in an average decrease of driving score by 18.87\%, compared to 6.78\% for other attack methods. In terms of the detailed metrics, our method performs slightly worse than ZOO-Adam \cite{chen2017zoo} in route completion (RC), but it achieves the lowest infraction score (IS), which typically results from serious hazardous behaviors such as running a red light, thereby leading to the lowest overall driving score. 
Additionally, due to the inherent instability in the driving simulation, the comparison of attack methods yields different results from the open-loop experiments, with the visual component of dual-modal attacks proving to be the most effective among the comparison methods.

In addition, we visualize the continuous frames of driving along Town 07 Route 79 in \Fref{fig:carla}, demonstrating the severe impact of our \tool attack. Under normal conditions, the driving agent can drive straight along the designated path. However, after applying our attack, the vehicle veers off the prescribed road to the left, colliding with roadside fences, which results in a high penalty to the final driving score.
More examples can be found in the Appendix.

\subsection{Real-world Experiments}
\label{sec:close-loop-real}

\begin{figure}[!t]
    \centering
    \includegraphics[width=0.98\linewidth]{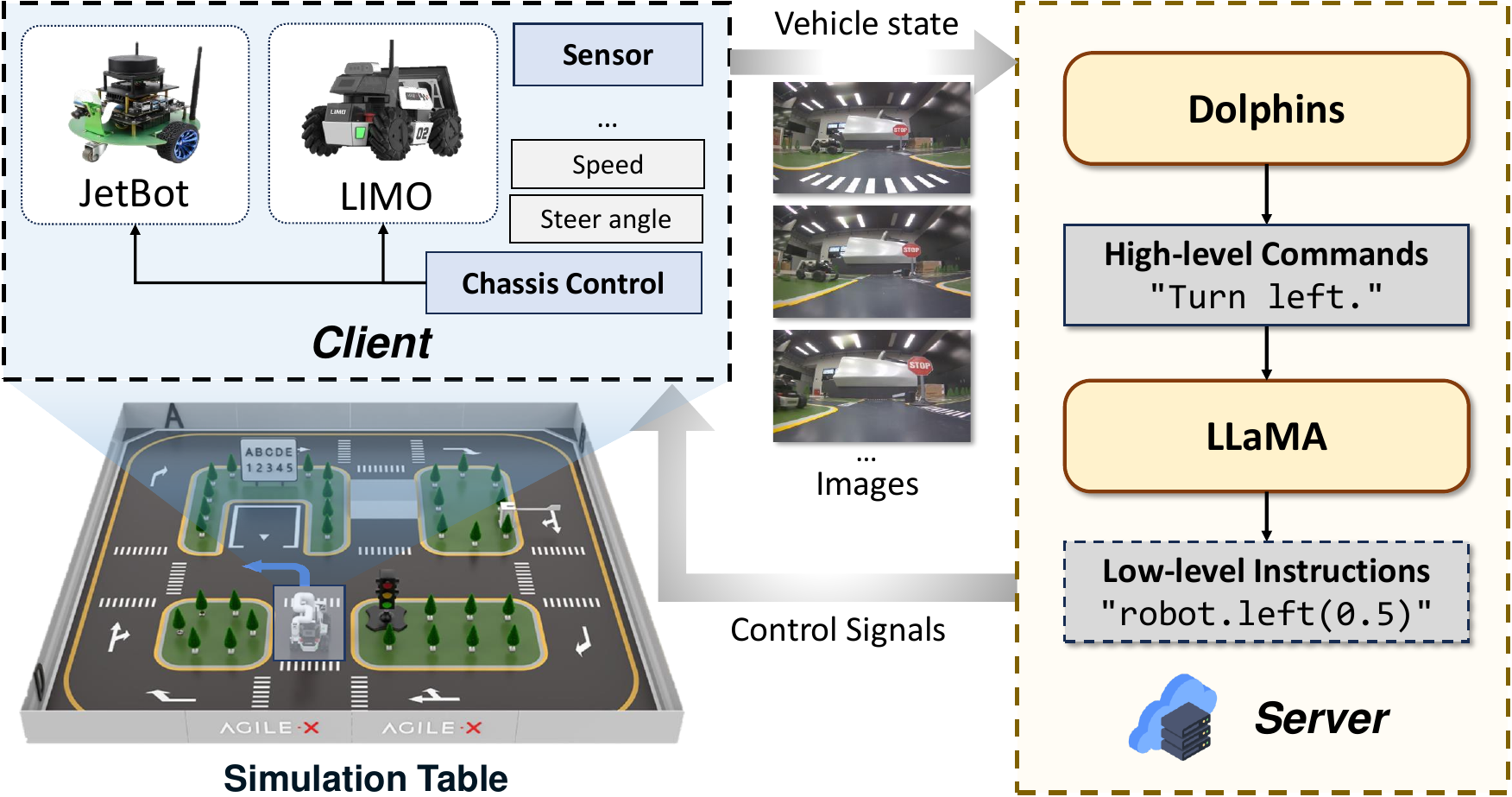}
    \caption{Overview of the VLM-driven pipeline. The system adopts a client-server architecture, using Dolphins to generate high-level commands and LLaMA to translate them into low-level control instructions.}
    \label{fig:driving_pipeline}
\end{figure}

In this section, we further investigate the adversarial robustness of AD systems driven by VLMs in the real world. The VLM-driven pipeline is illustrated in \Fref{fig:driving_pipeline}.

\textbf{Real-world Robotic Vehicles.} We select two commercially available robotic vehicles suitable for AD research and development, \ie, JetBot \cite{jetbot} and LIMO \cite{limo}, to conduct real-world attack experiments. Both vehicles are equipped with high-definition cameras, radar, IMU sensors, and basic motion capabilities. The JetBot vehicle \cite{jetbot} is equipped with more powerful computational resources, focusing on the application of artificial intelligence and deep learning algorithms within AD systems, and utilizes differential drive for motion. In contrast, the LIMO vehicle \cite{limo} is more oriented towards control-level applications, and we choose the Ackermann steering mode for its motion. The experimental environment consists of a manually constructed driving simulation track.

\textbf{AD Tasks Driven by VLMs.} Since both robotic vehicles do not inherently support VLM-driven decision-making and control, we configure AD tasks for the experiment using advanced VLMs. Specifically, we define the task of the vehicle as road-following with compliance to traffic rules, utilizing the driving-specific Dolphins \cite{ma2023dolphins} for decision-making. This model generates high-level commands such as ``\texttt{Keep going straight}'', which are then passed to LLaMA \cite{touvron2023llama}. The LLaMA model converts these high-level commands into control instructions, such as wheel speeds and steering angles, based on the car's motion model (\ie, differential drive or Ackermann steering). Both models require no additional training, relying instead on prompt-based techniques to accomplish the road-following task. The communication between the vehicles and models is facilitated through sockets. More detailed task setup is detailed in the Appendix. 

\begin{figure}[t]
    \centering
    
    \begin{subfigure}{0.49\linewidth}
        \centering
        \includegraphics[width=1.0\linewidth]{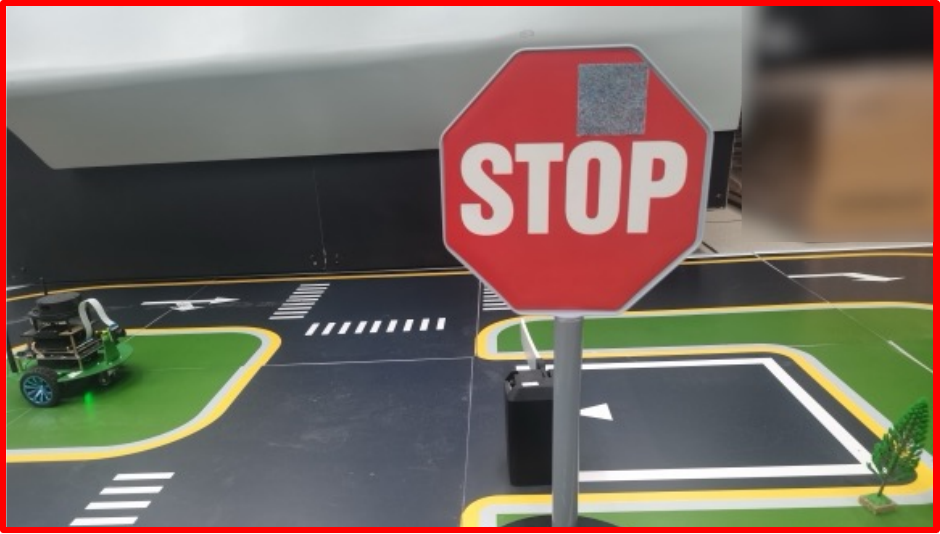}
        \caption{The ``STOP'' traffic sign .}
        \label{subfig:r2}
    \end{subfigure}
    \begin{subfigure}{0.49\linewidth}
        \centering
        \includegraphics[width=1.0\linewidth]{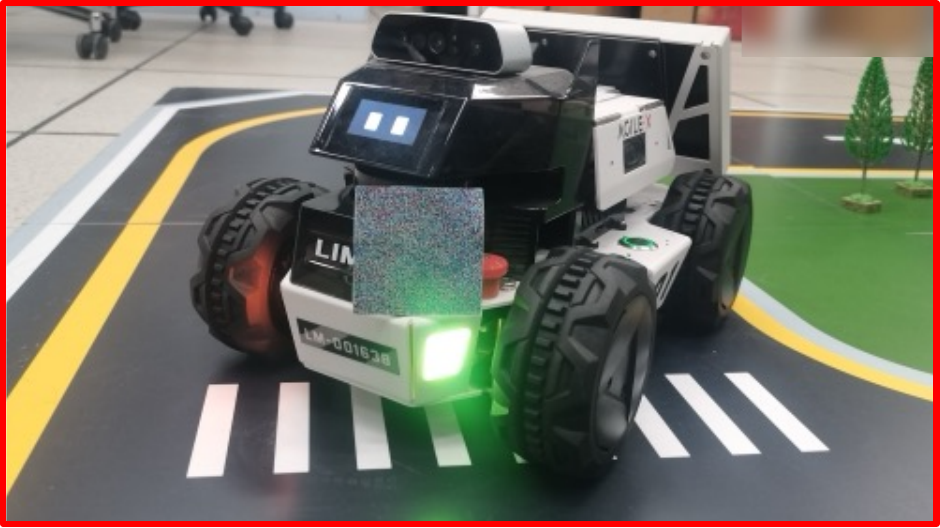}
        \caption{The obstacle vehicle.}
        \label{subfig:r4}
    \end{subfigure}
    
    \caption{Illustration of experimental scenarios and adversarial patches for Jetbot and LIMO in real-world settings.}
    \label{fig:real-world-attack}
\end{figure}

\textbf{Implementation of Attacks.} To evaluate the effectiveness of our method in real-world scenarios, we apply \tool in the form of adversarial patches \cite{liu2019perceptual}. Specifically, we utilize a commercially available 3D physical stop sign, along with another obstacle LIMO vehicle, as carriers for the adversarial patches. 
For generating the adversarial patches, we first capture images of the patch carriers on the simulation track respectively as the clean backgrounds. The adversarial patches, initialized randomly to 12\% of the original image size are then applied on the images. Subsequently, the patches are optimized without perturbation constraint according to the objective defined in \Sref{sec:overall}.
All other parameters remain consistent with those in \Sref{sec:setup}. Ultimately, the generated patches are printed and applied to the 3D traffic sign and obstacle vehicle, and we compare their effectiveness in the scenario with and without patches. Our experimental scenario and printed adversarial patches are shown in \Fref{fig:real-world-attack}.

\begin{table}[!t]
\caption{Experimental results in the real world. The values are shown in \emph{success times / total times}.}
\label{tab:real-exp}
\resizebox{\linewidth}{!}{
\centering
\begin{tabular}{@{}ccccccc@{}}
\toprule
 & \multicolumn{3}{c}{\textbf{JetBot} \cite{jetbot}} & \multicolumn{3}{c}{\textbf{LIMO} \cite{limo}} \\ \midrule
                     & Route 1 & Route 2 & Route 3 & Route 1 & Route 2 & Route 3 \\ \midrule
Origin               & 3 / 3   & 2 / 3   & 2 / 3   & 2 / 3   & 3 / 3   & 1/ 3    \\ \midrule
\tool & 1 / 3   & 0 / 3   & 1 / 3   & 0 / 3   & 0 / 3   & 0 / 3   \\ \bottomrule
\end{tabular}
}
\end{table}

\begin{figure}[!t]
    \centering
    \begin{subfigure}{0.98\linewidth}
        \centering
        \includegraphics[width=1.0\linewidth]{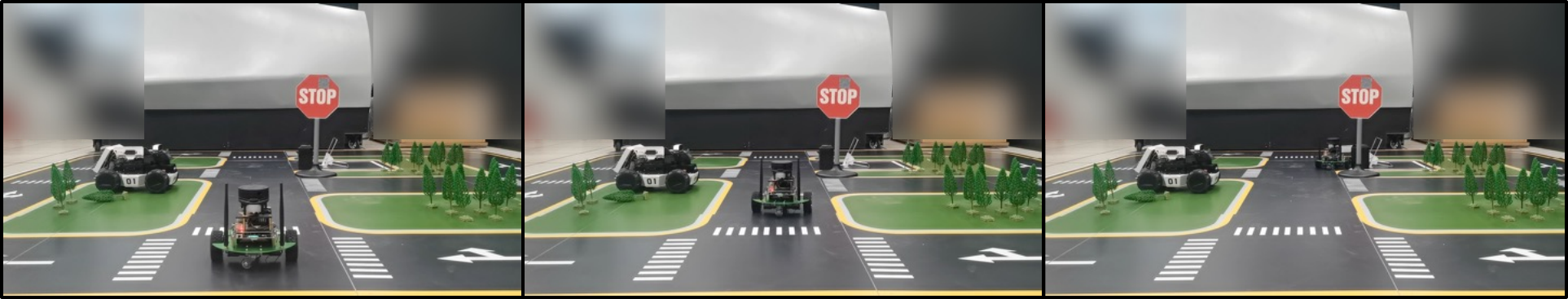}
        \caption{The Jetbot vehicle bypasses the stop sign.}
        \label{subfig:rr1}
    \end{subfigure}
    
    \begin{subfigure}{0.98\linewidth}
        \centering
        \includegraphics[width=1.0\linewidth]{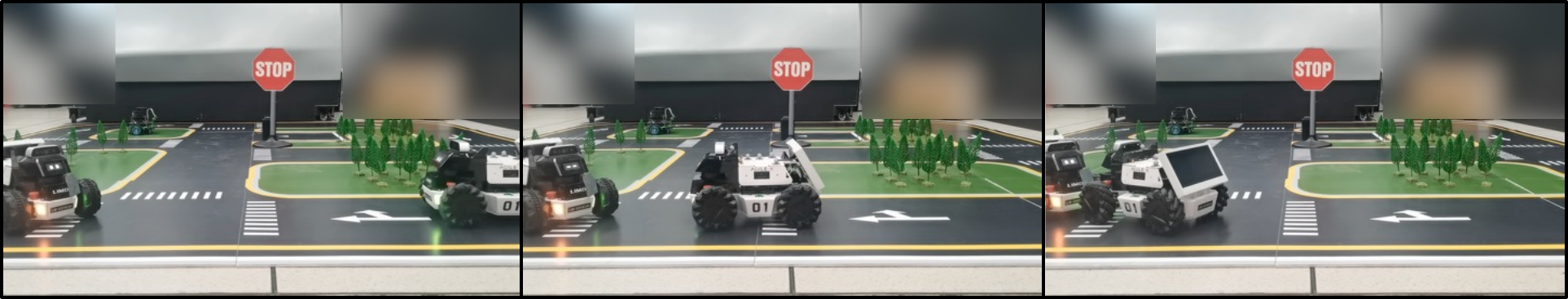}
        \caption{The LIMO vehicle crashes directly into the obstacle vehicle.}
        \label{subfig:rr2}
    \end{subfigure}
    
    \caption{Consecutive frames showing Jetbot and LIMO driven by VLMs, failing to interpret the required actions faced with adversarial patches, resulting in catastrophic collisions.}
    \label{fig:real-world-results}
\end{figure}

\textbf{Results Analysis.} We select three different driving routes for each of the two vehicles and perform both attack and non-attack runs. 
Each scenario is repeated three times, resulting in a total of 36 runs. 
We define a successful run as one where the vehicle completes the route without any collisions. The performance of the vehicles is presented in \Tref{tab:real-exp}. 
In the non-attack scenario, the successful completion rate of the two vehicles in our driving task reaches 72.22\%, demonstrating the effectiveness of our VLM-driven driving approach. 
Additionally, under normal conditions, the JetBot vehicle \cite{jetbot} exhibits smoother driving and a higher task completion rate than LIMO \cite{limo}. 
After applying the adversarial patches, the successful completion rate of two vehicles drops to 11.11\%, effectively validating the success of our attack in real-world applications. 
\Fref{fig:real-world-results} shows consecutive frames of the two VLM-driven vehicles exhibiting erroneous behavior when subjected to our attack. 
Both vehicles fail to correctly interpret the required actions, with the JetBot vehicle \cite{jetbot} bypassing the stop sign and the Limo vehicle \cite{limo} crashing directly into the obstacle vehicle with the adversarial patch.

\subsection{Ablation Studies}
\label{sec:ablation}

\begin{figure}[!t]   \centering\includegraphics[width=0.98\linewidth]{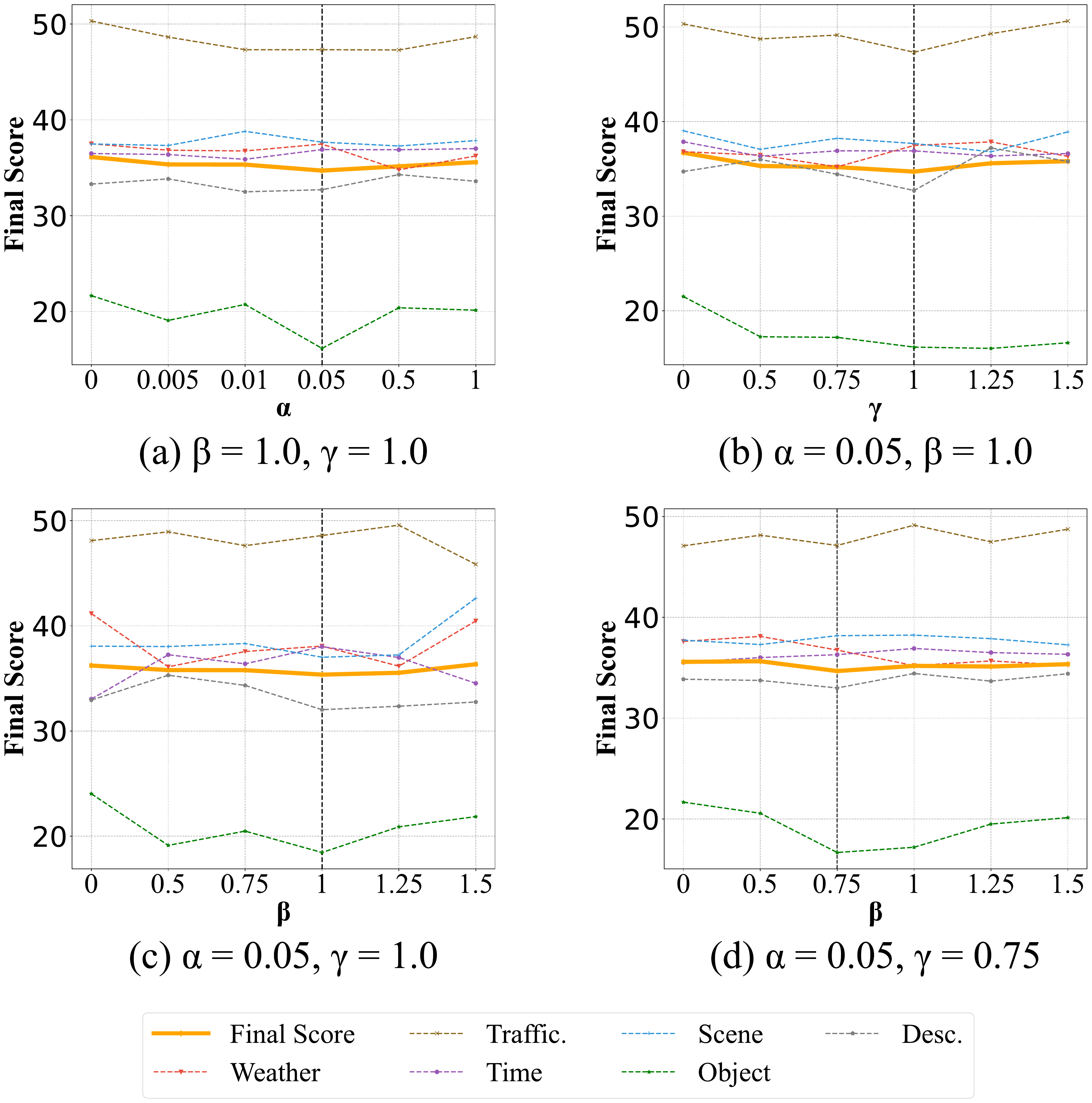}
    \caption{Ablation studies on decision chain disruption (reflected by $\alpha$), risky scene induction (reflected by $\beta$) and semantic discrepancy maximization (reflected by $\gamma$). The best experimental settings are marked with black dashed lines for each parameter adjustment. We use a controlled variable approach to adjust the next parameter under the two optimal settings of the present parameter.}
\label{fig:ablation-zhexian}
\end{figure}


\begin{figure}[!t]
\centering
	\begin{subfigure}{0.495\linewidth}
		\centering
        \includegraphics[width=0.98\linewidth]{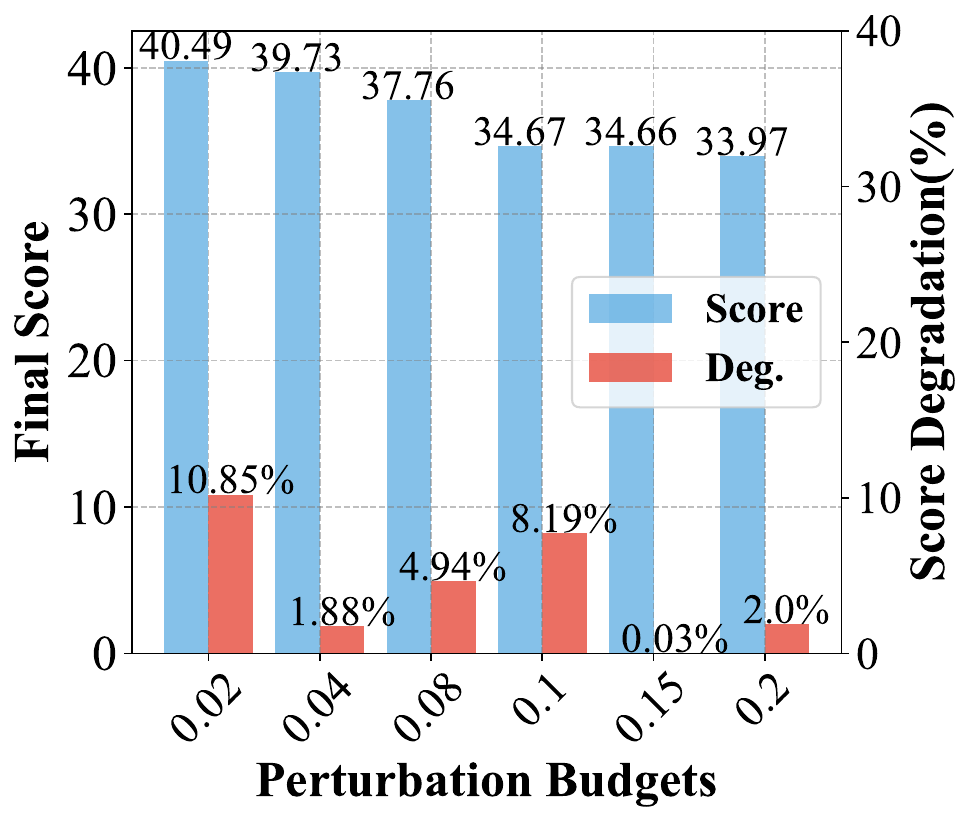}
		\caption{Different Budgets}
		\label{subfig:ablation-eps}
	\end{subfigure}
	\begin{subfigure}{0.495\linewidth}
		\centering
		\includegraphics[width=0.98\linewidth]{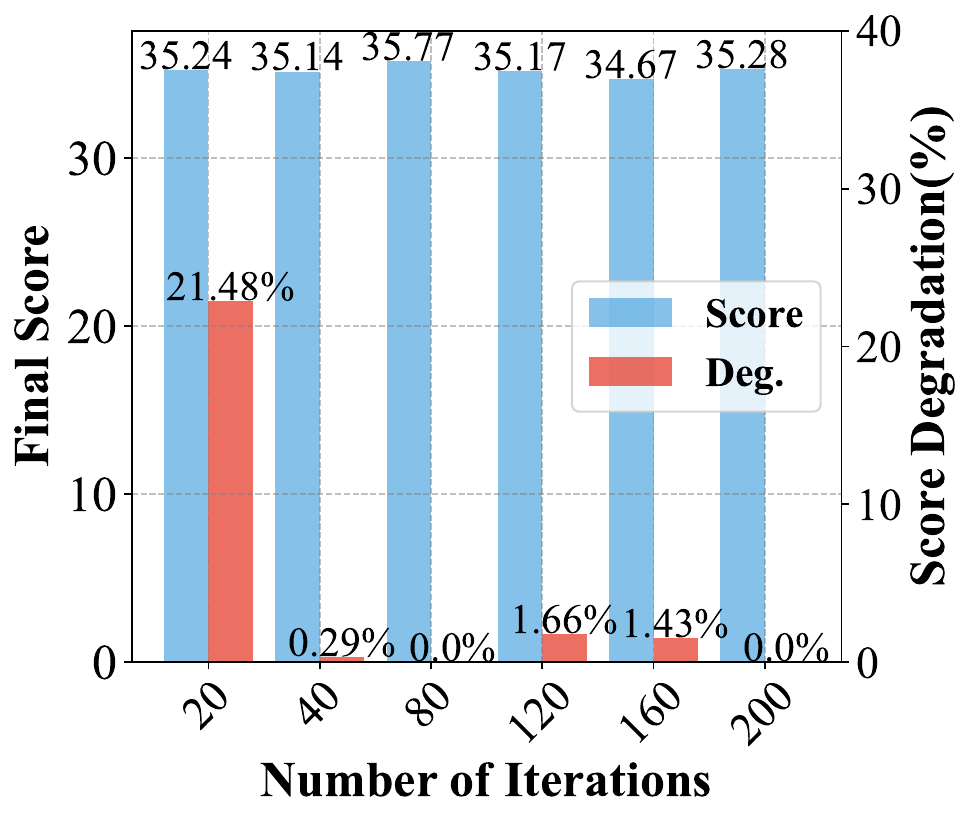}
		\caption{Different Iterations}
		\label{subfig:ablation-iter}
	\end{subfigure}
\caption{Ablation results on perturbation budgets and number of iterations. Score represents the final score of the model, while Deg. represents the percentage of performance degradation for each setting group compared to the previous one. If the degradation is negative, we set it to 0 directly in the graph.}
\label{fig:ablation-zhuzhuang}
\end{figure}

In this section, we investigate several key factors that might impact the performance of \toolns, thereby providing comprehensive insights and promoting a deeper understanding of our strategy. All the experiments conducted in this part use the Dolphins \cite{ma2023dolphins} target model on the Dolphins Benchmark.

\textbf{The key roles of decision chain disruption, risky scene induction and semantic discrepancy maximization.} 
To further analyze the contribution of each component in our attack design, we perform an ablation study by adjusting the corresponding hyperparameters for loss weight.
Specifically, the impact of decision chain disruption in \Sref{sec:dcd} and risky scene induction in \Sref{sec:rsi} is reflected by $\alpha$ and $\beta$, while that of semantic discrepancy maximization in \Sref{sec:overall} is reflected by $\gamma$.
We adopt a controlled variable approach. First, we set all three parameters to 1.0. Then, we sequentially adjust one parameter at a time to observe its impact on the attack results and identify the optimal configuration. Once the best two settings for the current ablated parameter are determined, we proceed with the ablation of the next parameter for each of these two best settings. Additionally, we also conduct complete ablation experiments for each component, which is represented by setting the corresponding parameter to 0. 

The ablation results are presented in \Fref{fig:ablation-zhexian}. In our progressive ablation experiments, it is evident that when a specific component (\ie, decision chain disruption, risky scene induction or semantic discrepancy maximization) is completely eliminated, the attack performance degrades significantly, with an average attack performance drop of 19.66\%, 13.94\% and 15.22\% respectively. 
Moreover, during the process of identifying the optimal settings, we find that the weight of each objective loss does not have a monotonic effect on the final results. Specifically, both excessively large or small weights lead to a reduction in attack effectiveness. 
Notably, when adjusting the parameter $\beta$, the optimal setting does not coincide with the best setting obtained from the previous parameter (\ie, $\alpha = 0.05$ and $\gamma = 1.0$), but rather corresponds to a suboptimal result (\ie, $\alpha = 0.05$ and $\gamma = 0.75$). To some extent, this highlights the stronger contribution of the \emph{high-level} risky scene induction compared to the semantic discrepancy maximization.

\begin{table}[!t]
\centering
\caption{Evaluation results of different modality-aligned models. \textbf{Bold text} indicates the method with the strongest attack effect in each column. \colorbox[gray]{0.9}{Gray cells} represent comprehensive evaluation metrics. For all metrics, lower values ($\textcolor{blue}{\downarrow}$) indicate stronger attack performance.}
\label{tab:abla-clip}

\resizebox{\linewidth}{!}{
\centering
\setlength{\tabcolsep}{1.8pt}
\centering
\begin{tabular}{@{}ccccccc>{\columncolor{gray!20}}cc@{}}
\toprule
\textbf{Method} & Weather$\textcolor{blue}{\downarrow}$ & Traffic$\textcolor{blue}{\downarrow}$ & Time$\textcolor{blue}{\downarrow}$ & Scene$\textcolor{blue}{\downarrow}$ & Object$\textcolor{blue}{\downarrow}$ & Desc.$\textcolor{blue}{\downarrow}$ & Final Score$\textcolor{blue}{\downarrow}$ \\ \midrule

ALBEF \cite{li2021align}                & 47.89 & 51.18 & 37.31 & 40.85 & 25.55 & 36.84 & 39.94  \\
VLMo \cite{bao2022vlmo}                      & 46.09 & 51.76 & 38.49 & 44.88 & 28.33 & 39.15 & 41.45  \\

CoCa \cite{yu2022coca}                 & 34.38 & 48.09 & \textbf{35.24} & \textbf{37.84} & 18.29 & 36.75 & 35.10  \\
BLIP \cite{li2022blip}                 & \textbf{28.39} & \textbf{42.88}  & 37.44 & 41.09  & 19.96  & 38.71 & 34.74 \\
\midrule
CLIP \cite{radford2021clip} & 36.75 & 47.13 & 36.30 & 38.18 & \textbf{16.67} & \textbf{32.99} & \textbf{34.67}  \\ \bottomrule
\end{tabular}
}

\end{table}

\textbf{The choice of pre-trained modality-aligned model.} 
To investigate the most effective pre-trained modality-aligned models for enhancing our attack, we conduct ablation experiments with five different state-of-the-art models: CLIP \cite{radford2021clip}, ALBEF \cite{li2021align}, VLMo \cite{bao2022vlmo}, CoCa \cite{yu2022coca}, and BLIP \cite{li2022blip}. All parameters are consistent with those in \Sref{sec:setup}. As shown in \Tref{tab:abla-clip}, CLIP \cite{radford2021clip} outperforms all other models, confirming its superior capability in efficiently aligning images and texts through contrastive learning. CoCa \cite{yu2022coca} and BLIP \cite{li2022blip} slightly lag behind CLIP \cite{radford2021clip}, while VLMo \cite{bao2022vlmo} and ALBEF \cite{li2021align} exhibit weaker performance in our attack task. These results highlight the effectiveness of CLIP's simple yet powerful modality alignment ability, validating its selection as the pre-trained model for our framework.

\textbf{Perturbation budgets and number of iterations.}
We evaluate our attack across various perturbation budgets (\ie, $\ell_\infty$ norm values of 0.02, 0.04, 0.08, 0.1, 0.15, 0.2 with steps = 160) and different iterations (\ie, 20, 40, 80, 120, 160, 200 with $\epsilon$ = 0.1). The results are presented in \Fref{fig:ablation-zhuzhuang}. We report the final score for each configuration, along with the performance drop compared to the previous setting. 
A larger perturbation budget leads to a lower score, which aligns with intuitive expectations. Notably, when $\epsilon$ is set to 0.1, the highest percentage performance drop (8.19\%) is observed compared to the previous configuration across all adjustments. Regarding the number of iterations, the results do not follow a strictly consistent pattern, but generally, a larger number of iterations tends to produce stronger attack effects.

\section{Countermeasures against \toolns}

\begin{table}[!t]
\footnotesize
\centering

\caption{Countermeasure studies on three AD VLMs. \textbf{Bold text} indicates the method with the strongest defense effect in each column. We report the final results using the corresponding metrics for each model, higher values ($\textcolor{red}{\uparrow}$) indicate stronger defense performance. 
}
\label{tab:defense}
\setlength{\tabcolsep}{2.1pt}
\resizebox{\linewidth}{!}{
\centering
\begin{tabular}{@{}ccccc}
\toprule
\multicolumn{1}{c}{\textbf{Categoty}}                                               & \textbf{Method} & \textbf{Dolphins} \cite{ma2023dolphins}$\textcolor{red}{\uparrow}$ & \textbf{DriveLM} \cite{sima2023drivelm}$\textcolor{red}{\uparrow}$ & \textbf{LMDrive} \cite{shao2024lmdrive}$\textcolor{red}{\uparrow}$ \\ \midrule
\multicolumn{1}{c}{}                         & JPEG         & 31.80   & 44.76 & 42.35                  \\
\multicolumn{1}{c}{}                         & TVM          & 26.05  & 44.24 & 43.33                   \\
\multicolumn{1}{c}{\multirow{-3}{*}{I.T \cite{guo2017countering}}}                        & MS              & 34.95 & 47.90 & 46.62                         \\ \midrule
\multicolumn{1}{c}{A.D \cite{xu2017feature}}    & Bit-Red      & 35.17 &        47.92       &  45.28   \\ \midrule
\multicolumn{1}{c}{I.D \cite{naseer2020self}}          & NRP          & \textbf{39.61} & 46.74 & \textbf{47.47} \\ \midrule
\multicolumn{1}{c}{O.P \cite{chen2022adversarial}}  & Post-Process & 35.21  & \textbf{48.03} & 41.38                  \\ \midrule
\multicolumn{1}{c}{} & Text-Chain    &  35.28  & 46.86 & 42.00                 \\
\multicolumn{1}{c}{\multirow{-2}{*}{T.E}} &  Text-Scene      & 35.60   & 46.75 & 44.53                    \\ \midrule
\multicolumn{2}{c}{No defense}                                     & 34.67 & 46.32 & 39.99                      \\ \bottomrule
\end{tabular}

}

\end{table}

In this section, we explore defense strategies to mitigate the potential negative social impacts. We incorporate several well-established defense techniques from input pre-processing to output post-processing; additionally, we design a textual enhancement defense strategy to counteract our attack. The attack settings are consistent with those outlined in \Sref{sec:setup}. 

\textbf{Image Transformation (I.T)} applies a series of image transformation operations prior to model input \cite{guo2017countering}, aiming to eliminate or diminish the finely crafted perturbations introduced by adversarial attacks, thereby significantly reducing their disruptive effects on the model. Specifically, we select three classic transformation operations for this defense approach, including JPEG compression \cite{guo2017countering}, median smoothing \cite{chiang2020detection}, and TVM (Total Variation Minimization) \cite{guo2017countering}.

\textbf{Adversarial Detection (A.D)} mitigates adversarial attacks by identifying inputs with anomalous features indicative of adversarial manipulation. If an adversarial sample is detected, the system either rejects the input or applies corrective measures. Specifically, we use the bit-depth compression method outlined in \cite{xu2017feature} for detection.
Detected adversarial inputs are replaced with clean counterparts.

\textbf{Image Denoising (I.T)} typically introduces pre-processors before model inference to remove adversarial noise or use image reconstruction techniques to eliminate adversarial features within the image. Specifically, we employ NRP (Neural Representation Purification) \cite{naseer2020self} to denoise the input image before it is fed into the AD VLMs for inference.

\textbf{Output Post-processing (O.P)} focuses on post-processing the model’s output to purify it. Inspired by AAA \cite{chen2022adversarial}, which proposes post-processing to counter black-box query attacks, we apply this concept to defend against adversarial attacks on AD VLMs. Specifically, we employ rule-based filtering and correction of AD VLMs’ outputs. For instance, in the VQA tasks on Dolphins \cite{ma2023dolphins} and DriveLM \cite{sima2023drivelm}, we search for extreme driving behaviors in the output and replace them with safe driving actions such as maintaining a smooth driving trajectory. In the case of the closed-loop LMDrive \cite{shao2024lmdrive}, we directly detect extreme control commands (\eg, steering angles) and constrain them to safer and smoother ranges.

\textbf{Textual Enhancement (T.E)}. The aforementioned defense methods primarily focus on mitigating attacks through the visual domain at the input level or via output filtering, yet they lack defenses targeting the textual domain at the input level. 
To provide a more comprehensive defense evaluation, we propose Textual Enhancement as a new defense mechanism.
We aim to design a defense strategy that directly targets our attack methodology. 
Drawing inspiration from the text-related design, we incorporate beneficial prompts into the textual input to weaken the corresponding disruptive semantics. 
Specifically, for \emph{low-level} Decision Chain Disruption, we emphasize the logical consistency of the perception, prediction, and planning chain in the original prompt (\eg, accurately identify potential unexpected obstacles in the environment and their future state) to construct advantageous cues. This approach helps weaken the deception that does not align with normal driving reasoning. 
For \emph{high-level} Risky Scene Induction, we incorporate safety-related constraints and guidelines (\eg, avoid collision risks) into the original prompt to formulate beneficial cues. These cues are intended to guide the model towards making decisions that prioritize holistic safety rather than merely focusing on driving actions as a measure of safety. We refer to them as \emph{Text-Chain} and \emph{Text-Scene} respectively.

\textbf{Results and Discussion.}
Despite the significant challenges posed by our attack, the proposed defense strategies are still able to mitigate its negative impact to some extent, as illustrated in \Tref{tab:defense}. The breakdown results can be found in the Appendix. Specifically, the denoising method stands out as the most effective among all the defenses, demonstrating performance improvements of 14.25\%, 0.91\%, and 18.70\% across the three models when subjected to the \tool Attack. Among the other defense approaches, image transformation-based methods perform the weakest, with some even exhibiting a reverse defense effect, resulting in worse performance after defense implementation. We attribute this to the fact that while image transformation operations alleviate the interference from adversarial perturbations, they also introduce additional loss of image information, which compromises the overall performance. In contrast, detection and post-processing-based defenses exhibit more promising results, highlighting their superior ability to counter adversarial attacks. Furthermore, the textual enhancement we proposed also provides a degree of defense, especially the \emph{Text-Scene}, but overall, its effectiveness remains inferior to that of denoising methods. This suggests that more sophisticated mechanisms need to be developed to effectively counter our attacks. Note that, due to the high computational overhead and the sparsity of literature in the VLM area,  we do not include adversarial training as a defense in our paper. We leave it as the future work.

\section{\data Dataset}

Datasets play a crucial role in advancing model research, particularly in areas where specialized benchmarks are lacking or where data collection is prohibitively expensive. In light of this, we propose a safety evaluation dataset specifically for AD VLMs, aimed at fostering the development of more resilient AD VLMs.

\subsection{Construction Details}

Our \tool Attack dataset (\data) includes Scene-\data and Obj-\datans, which correspond to scene-level and object-level adversarial perturbations, respectively.

\textbf{Data collection.} 
For Scene-\data, we mainly consider injecting adversarial perturbations on the driving scenes (\ie, videos and images). Specifically, we adopt our \tool on visual images/videos in Dolphins Benchmark \cite{ma2023dolphins} and DriveLM-nuScenes \cite{sima2023drivelm} datasets. In contrast to our main experiments, during the noise generation process, we increase the number of queries to the auxiliary VLM to five, utilizing the aggregated results from all queries during optimization iterations in Decision Chain Disruption; we extend the set of opposite safety descriptors to five distinct groups in the Risky Scene Induction, enabling a broader basis for subsequent matching tasks. This augmentation enriches the diversity and complexity of the adversarial inputs, ensuring a more robust evaluation of model vulnerabilities. 
To construct a more practical dataset, we design Obj-\data composed of a set of traffic signs with adversarial patches inspired by the real-world experiments in \Sref{sec:close-loop-real}. Specifically, we collect a set of 140 traffic sign images, each with a size of 224 * 224, from publicly available resources on the Internet. The images include both pure logo images and real-world road images, covering common directive signs, prohibition signs, and warning signs. Using these images as the raw dataset, we apply the \tool Attack to generate adversarial patches. The attack settings follow the same configuration as those described in \Sref{sec:close-loop-real}, and we attach them to the corresponding clean  images to create printable adversarial traffic signs.

\begin{figure}[!t]
    \centering
    \includegraphics[width=0.98\linewidth]{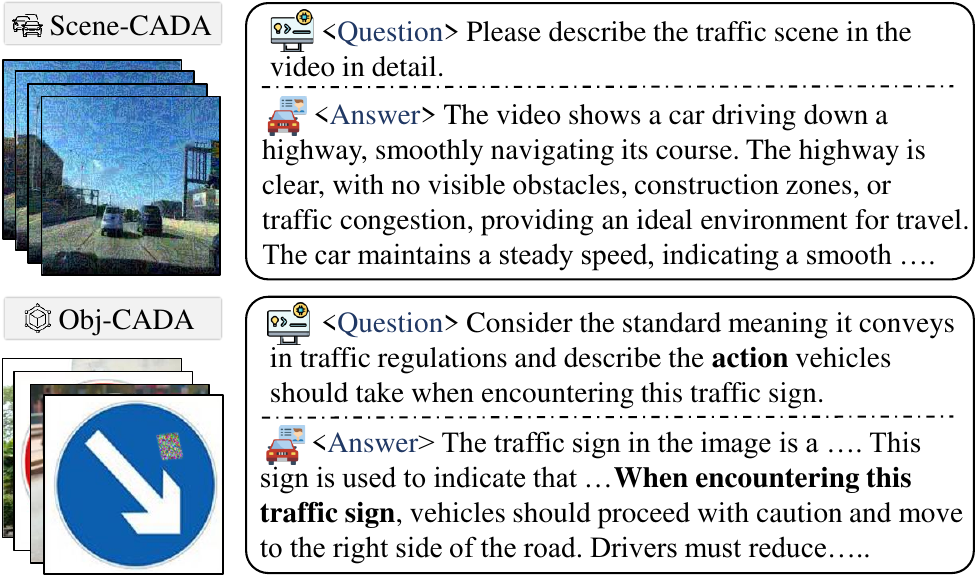}
    \caption{Examples of our \data Dataset, which correspond to scene-level and object-level perturbations, respectively.}
    \label{fig:dataset}
\end{figure}

\textbf{VQA pairs construction.} For Scene-\data, we match each set of noisy videos or images with the original QA pairs from the Dolphins Benchmark \cite{ma2023dolphins} and DriveLM \cite{sima2023drivelm} datasets. 
For Obj-\data, we design and generate new corresponding QA pairs to facilitate testing within AD tasks. In contrast, the questions focus on the appropriate actions a vehicle should take in response to the traffic signs rather than explaining the meaning of traffic signs only, and the answers specify the corresponding vehicle behaviors. The illustrations can be found in \Fref{fig:dataset}.

\textbf{Quality control procedure.} 
We adopt a quality control procedure similar to those employed by many VQA datasets \cite{sima2023drivelm, xu2024drivegpt4} for Obj-\data. Leveraging GPT \cite{chatgpt}, we generate three sets of QA pairs with varying expression forms per visual data instance using a structured template. Human expert annotators then follow consistent guidelines to evaluate these pairs and refine them through corrections, finally selecting the best one for inclusion.
To ensure accuracy, annotators are divided into three groups. Each visual question-answer pair is randomly assigned to two groups for initial review, with a final group designated for confirmation.

\subsection{Data Properties}

\quad\textbf{Severity ranking.} 
Our dataset is systematically stratified into 4 distinct severity levels (0 to 4, where ``0'' denotes clean). This results in 4 times the amount of adversarial data compared to the original clean dataset, thereby enabling a more comprehensive evaluation under varying intensities of adversarial perturbations. 
Specifically, for scene-\data, we introduce 4 different perturbation magnitudes in terms of the $\ell_\infty$ norm: 0.02, 0.04, 0.06, and 0.08, corresponding to 4 severities. 
For Obj-\data, we categorize 4 severity levels based on the size of adversarial patches, corresponding to 10\%, 15\%, 20\%, and 25\% of the original side length of the traffic sign images. Each patch is randomly placed within the image.

\textbf{Dataset scale.}
Our Scene-\data comprises 486 clean and 1,944 adversarial video-question-answer pairs as well as 4,076 clean and 16,304 adversarial image-question-answer pairs, encompassing key tasks of autonomous driving, including scene description, traffic object recognition, motion prediction, decision planning, and behavior explanation. 
Obj-\data consists of 140 clean and 560 adversarial traffic sign image-question-answer pairs, and the task mainly focuses on analyzing driving actions constrained by traffic sign interpretations. 
Overall, our dataset contains 2,430 video pairs and 21,080 image QA pairs, including 18,808 adversarial visual QA pairs.


\subsection{Preliminary Experiments}

\begin{table}[!t]
\footnotesize
\centering

\caption{Evaluation results on \data dataset. We report the final results using the corresponding metrics for each model, higher values ($\textcolor{red}{\uparrow}$) indicate stronger 
model robustness.}
\label{tab:dataset-exp}
\setlength{\tabcolsep}{1.5pt}
\resizebox{\linewidth}{!}{
\centering
\begin{tabular}{@{}ccccccc}
\toprule
\multicolumn{1}{c}{\textbf{Dataset}}                                               & \textbf{Model} & \textbf{Level 0}$\textcolor{red}{\uparrow}$ & \textbf{Level 1}$\textcolor{red}{\uparrow}$ & \textbf{Level 2}$\textcolor{red}{\uparrow}$ & \textbf{Level 3}$\textcolor{red}{\uparrow}$ & \textbf{Level 4}$\textcolor{red}{\uparrow}$  \\ \midrule
\multicolumn{1}{c}{}                         & Dolphins \cite{ma2023dolphins}          & 44.88  & 40.48 & 39.73 & 37.76 & 35.57  \\
\multicolumn{1}{c}{\multirow{-2}{*}{Scene-\data}} & DriveLM \cite{sima2023drivelm} & 53.55 &  48.16 & 47.36 & 47.67 & 46.96 \\ \midrule 
\multicolumn{1}{c}{Obj-\data}    & Dolphins \cite{ma2023dolphins} & 56.56 &  54.17  & 50.80 & 47.81 & 46.54 \\ 
\bottomrule
\end{tabular}
}

\end{table}


Here, we further evaluate AD VLMs \cite{ma2023dolphins} on our dataset, where Dolphins \cite{ma2023dolphins} and DriveLM \cite{sima2023drivelm} are evaluated on their respective Scene-\data subsets. Additionally, Dolphins is also evaluated on Obj-\data through GPT Score (higher the better performance) due to its superior zero-shot capability.
As shown in \Tref{tab:dataset-exp}, we can identify that  \ding{182} Both our Scene-\data and Obj-\data present significant threats to AD VLMs, leading to a max performance degradation of 16.53\% and 17.72\%, respectively. \ding{183} As the severity of adversarial perturbations increases, both models exhibit a noticeable performance drop trend, with the average model performance at level 4 being 9.57\% lower than that at level 1. \ding{184} The performance degradation caused by Obj-\data is higher than that caused by Scene-\data, highlighting the importance of prioritizing defenses against physical attacks in real-world applications.

These findings highlight the necessity for more advanced training strategies and robust defense mechanisms. We encourage future research to assess their effectiveness based on our \data dataset.

\section{Related Work}

Adversarial attacks, first introduced by Szegedy \etal \cite{szegedy2013intriguing}, are inputs designed to mislead deep learning models, often visually imperceptible to humans. Initially, adversarial attacks often target visual models to induce misclassification or other erroneous predictions \cite{goodfellow2014fgsm, PGD, liu2019perceptual, liu2020bias, carlini2017towards}. Extensive research has refined attack strategies, including gradient-based methods such as FGSM \cite{goodfellow2014fgsm} and PGD \cite{PGD}, as well as perceptual attacks designed to disrupt semantic consistency \cite{liu2019perceptual, liu2020bias, liu2022harnessing, liu2020spatiotemporal, wang2021dual}. Generally, adversarial attacks can be classified into \textit{white-box} and \textit{black-box} settings based on the attacker's capabilities. In a \textit{white-box} setting, the adversary has access to the model's architecture and gradients, enabling the design of more targeted attacks. Conversely, in a \textit{black-box} setting, perturbations must be crafted solely based on observable model outputs without knowledge of the model's internal workings. 


In the context of classical autonomous driving, adversarial attacks have been applied to exploit vulnerabilities in perception modules critical to autonomous driving. For example, Wang \etal \cite{wang2021dual} proposed an adversarial attack targeting vehicle detection systems, demonstrating how adversarial camouflage can compromise safety-critical vehicle recognition. Similarly, Sato \etal \cite{sato2021dirty} introduced adversarial patterns designed to mislead lane detection systems, impairing a vehicle's ability to maintain its trajectory. However, these attacks focus exclusively on traditional models and do not address the emerging challenges VLM-based models pose.
\textbf{In the context of AD VLMs}, Zhang \etal \cite{zhang2024visual} introduced ADvLM, the first white-box adversarial attack specifically targeting AD VLMs. This work investigates the unique characteristics of AD scenarios by analyzing attack sources and expanding the potential attack space to include both texts and images, highlighting vulnerabilities inherent in AD VLM systems.

Our paper \textbf{differs} from previous studies, as ADvLM \cite{zhang2024visual}, the only adversarial attack in the field of AD VLMs, is specially designed for white-box attacks that necessitate the target model's gradients.
This paper presents the first specially designed \textbf{black-box adversarial attacks on AD VLMs}, requiring no gradient access to the target models.
\section{Conclusion and Future Work}

In this paper, we propose the \tool, the first black-box adversarial attack specifically tailored for VLM AD based on decision chain disruption and risky scene induction. We conduct extensive experiments on both AD VLMs and general VLMs, as well as the real AD vehicles driven by VLMs, showing the superiority of our attack. In addition, we present the \data dataset, comprising 18,808 adversarial visual-question-answer pairs, to facilitate research.

\textbf{Limitation and Future Work.} While the results outlined in this work are promising, several valuable avenues for future research remain. 
\ding{182} We would like to explore the textual domain for a more comprehensive attack framework.
\ding{183} We would like to incorporate visually inconspicuous attack design such as camouflage.
\ding{184} Future work would like to validate its attack against commercial AD systems.

\bibliographystyle{plain}
\bibliography{ref}



\end{document}